\documentclass[journal]{IEEEtran}
\usepackage{etoolbox}
\makeatletter
\patchcmd{\@makecaption}
  {\scshape}
  {}
  {}
  {}
\makeatother
\usepackage{amsmath,amssymb}
\usepackage{amsfonts}
\usepackage{tikz}
\usepackage{float}
\usepackage{graphicx}
\usepackage{pgfplots}
\ifCLASSINFOpdf

\else

\fi

\begin{document}

\title{Increased-Range Unsupervised Monocular Depth Estimation}
\author{\IEEEauthorblockN{Saad Imran\IEEEauthorrefmark{1},
Muhammad Umar Karim Khan\IEEEauthorrefmark{2}, Sikander Bin Mukarram\IEEEauthorrefmark{3} and
Chong-Min Kyung\IEEEauthorrefmark{4}}
\IEEEauthorblockA{\IEEEauthorrefmark{1},\IEEEauthorrefmark{3}Department of Electrical Engineering, Korea Advanced Institute of Science and Technology, Daejeon, 34141, South Korea\\
\IEEEauthorrefmark{2},\IEEEauthorrefmark{4}Center of Integrated Smart Sensors, Korea Advanced Institute of Science and Technology, Daejeon, 34141, South Korea\\
Email: \IEEEauthorrefmark{1}saadimran@kaist.ac.kr,
\IEEEauthorrefmark{2}umar@kaist.ac.kr,
\IEEEauthorrefmark{3}sikander\_ssalab@kaist.ac.kr,
\IEEEauthorrefmark{4}kyung@kaist.ac.kr}}

\maketitle
\begin{abstract}
Unsupervised deep learning methods have shown promising performance for single-image depth estimation. Since most of these methods use binocular stereo pairs for self-supervision, the depth range is generally limited. Small-baseline stereo pairs provide small depth range but handle occlusions well. On the other hand, stereo images acquired with a wide-baseline rig cause occlusions-related errors in the near range but estimate depth well in the far range. In this work, we propose to integrate the advantages of the small and wide baselines. By training the network using three horizontally aligned views, we obtain accurate depth predictions for both close and far ranges. Our strategy allows to infer multi-baseline depth from a single image. This is unlike previous multi-baseline systems which employ more than two cameras. The qualitative and quantitative results show the superior performance of multi-baseline approach over previous stereo-based monocular methods. For 0.1 to 80 meters depth range, our approach decreases the absolute relative error of depth by 24\% compared to Monodepth2. Our approach provides 21 frames per second on a single Nvidia1080 GPU, making it useful for practical applications.
\end{abstract}

\section{Introduction}
Depth estimation is a commonly studied problem in computer vision due to the large number of applications.  Accurate depth information is important for tasks such as 3D reconstruction and autonomous navigation. In the past, many methods such as time-of-flight \cite{zhu2008fusion}, structured light \cite{scharstein2003high}, stereo matching \cite{scharstein2002taxonomy}, and structure from motion (SFM) \cite{sturm1996factorization} have been used for depth estimation. 

Recently, with the advancement of deep learning, many researchers have used Convolutional Neural Networks (CNN) with self-supervision for single image depth estimation. Self-supervised methods for monocular depth estimation have shown promising performance in recent years. These methods treat depth estimation as an image reconstruction problem. More specifically, a CNN is trained to generate disparity maps that are used to reconstruct the target images from the reference images. Self-supervised methods are preferred over supervised deep learning approaches as the former do not require ground truth depth data, which is expensive and hard to gather.

Existing monocular depth estimation models can be trained by either using monocular video or rectified stereo pairs. These approaches have some challenges. Besides training a depth estimation network, monocular video-based methods also require relative pose information between the adjacent frames in a sequence. On the other hand, stereo approaches do not require training a pose estimation network and are more effective than video-based methods. Despite this advantage, existing stereo techniques use two cameras (left and right images) with a fixed baseline for training, which can cause occlusion and limited depth range.

To deal with these problems, researchers have proposed multi-camera systems \cite{honegger2017embedded,okutomi1993multiple} with multiple baselines. The advantage of a multi-baseline setup is that it can provide good depth accuracy both in near and far ranges compared to standard stereo, which only works well in a certain range. However, the problems with such systems are that they are quite expensive and have high computational load due to multiple cameras; hence, they are not commonly used. In this paper, we aim to improve disparity estimation with a monocular camera by leveraging multiple baselines at training time.

We present an unsupervised learning approach that uses two different baselines during training and a single image at test time. Our approach makes use of the advantages of multi-baseline stereo without increasing the computational complexity at inference. In contrast to two-camera stereo-based monocular methods, our method gives improved disparity maps in both near and far ranges. To our knowledge, we are the first to employ the principle of multi-baseline to unsupervised disparity estimation. Experimental results show that our method yields much improved results compared to stereo-based self-supervised monocular depth extraction.

\section{Related Work}
Although various methods have been proposed in the past to extract depth from images, we discuss the literature related to our work only.

\textbf{Classical Stereo Depth Estimation.} Stereo matching that involves two horizontally displaced cameras to observe a given scene is one of the most popular approaches . The shift between the corresponding pixels in observed left and right images gives the disparity, which is inversely proportional to the depth at the pixels. Traditional stereo matching algorithms usually include all or some of the four tasks: computing the matching cost, aggregating the cost, computing the disparity, and refining the disparity. A detailed description of these tasks is given in \cite{scharstein2002taxonomy}. Among different stereo matching techniques, semi-global matching (SGM) \cite{hirschmuller2005accurate} is one of the most frequently used approaches because of its efficiency. Single baseline stereo setup poses some problems. For example, using wider baseline increases the chance of false matches due to large disparity search range. On the other hand, short baseline reduces the risk of false matching but suffer from poor accuracy in the far range. It is well known that using more views can solve these problems. \cite{ito1986three} proposed using three views in a triangular configuration to improve the matching and handle the occlusion. \cite{okutomi1993multiple} generated multiple baselines by laterally displacing the camera. They showed that matching across different baseline stereo images circumvents the problem of incorrect matching and results in more accurate disparity maps. \cite{honegger2017embedded} used four cameras in a parallel arrangement to propose a multi-baseline stereo system that works in real time. \cite{gallup2008variable} proposed a multi-baseline, multi-resolution technique and achieved a constant depth accuracy by changing the baseline and resolution accordingly.

\textbf{Supervised Monocular and Stereo Depth Estimation.} Many supervised learning based schemes have been proposed for depth extraction. Make3D \cite{saxena2008make3d} modified a Markov Random Field (MRF) to predict  the 3D structure of a scene from a single image. \cite{eigen2014depth} used two neural networks to infer monocular depth by combining global and local structure of the input image. \cite{liu2015learning} showed the advantage of jointly training a CNN and  a conditional random field (CRF) for monocular depth estimation. In \cite{fu2018deep}, authors treat monocular depth estimation as an ordinal regression problem, while in \cite{lee2019monocular}, authors use relative depth maps to achieve state-of-the-art results. Many researchers have used deep learning for stereo depth estimation. Zbontar and LeCun \cite{vzbontar2016stereo} proposed two network architectures to compare image patches for computing the stereo matching cost followed by traditional post-processing steps. \cite{luo2016efficient} obtained better results by treating stereo matching problem as a multi-class classification task. \cite{kendall2017end} presented an end-to-end learning technique for disparity estimation, which does not require additional post-processing. In \cite{yadati2017multiscale}, authors proposed a multi-scale approach to predict depth from unrectified stereo images. \cite{tosi2018beyond} combined local and global features to obtain accurate confidence scores. \cite{tulyakov2018practical} proposed a stereo matching technique to deal with large memory and dynamic disparity range requirements. \cite{zhang2019ga} introduced two additional layers for efficient and accurate cost aggregation. In \cite{duggal2019deeppruner}, authors proposed a faster method of stereo matching by discarding most of the disparities during cost aggregation. 

\textbf{Unsupervised Monocular and Stereo Depth Estimation.} Currently, unsupervised methods for depth estimation are becoming more popular as they do not require expensive ground truth data. Garg \textit{et al.} \cite{garg2016unsupervised} were the first to propose a fully unsupervised approach for monocular depth estimation. They used rectified stereo images for training and performed Taylor series expansion to linearize the image warping process. \cite{godard2017unsupervised} used the bilinear sampler \cite{jaderberg2015spatial} for image warping and introduced left-right consistency loss to obtain accurate depth results. In \cite{poggi2018learning}, authors imposed trinocular stereo assumptions to yield enhanced depth results. \cite{poggi2018towards} deployed a pyramidal architecture to enable monocular depth estimation on embedded systems, while the authors in \cite{pilzer2018unsupervised} used adversarial learning for stereo depth estimation. \cite{tosi2019learning} added proxy supervision by obtaining disparity maps through traditional stereo matching technique. SuperDepth \cite{pillai2019superdepth} incorporated Single-Image Super-Resolution (SISR) \cite{shi2016real} technique to obtain high resolution disparity maps. Recently, many researchers have used monocular videos for self-supervision to predict depth from a single image \cite{godard2019digging,gordon2019depth,zhou2019unsupervised}.

\section{Unsupervised Multi-Baseline approach}
\begin{figure}[t!]
      \centering
     \includegraphics[width=0.3\textwidth]{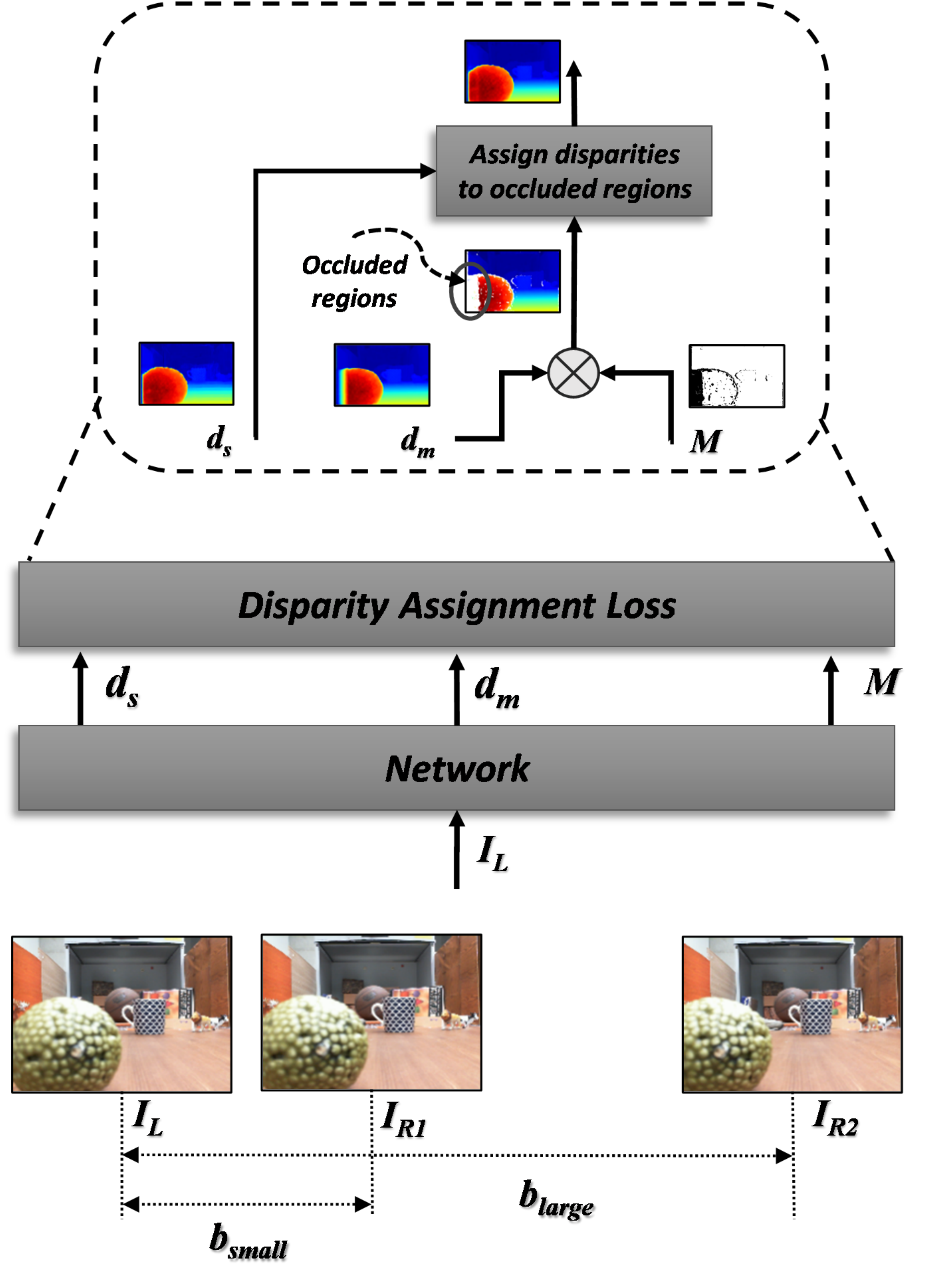}
      \caption{The concept behind our training approach. $b_{small}$ and $b_{large}$ refer to small and wide baselines, respectively.}

    \label{fig:concept}
\end{figure}

Although supervised learning can be used for multi-baseline stereo, it has its problems. We can extend the work of \cite{tulyakov2018practical,vzbontar2016stereo} to match more than two views, however, using such approaches have two major drawbacks. First, they require pixel-wise labelling to generate ground truth depths for training. Second, we also need to use more than two cameras at test time to perform matching, which is highly undesirable. Therefore, we train the model in a self-supervised fashion and use a single image for inference.

Figure \ref{fig:concept} shows the basic idea of our approach. For training, we use three aligned views to get two different baselines. The images $I_{L}$ and $I_{R1}$ act as a small-baseline stereo system, and hence have fewer occlusions \cite{delon2007small} but provide accurate near depth. On the other hand, the images $I_{L}$ and $I_{R2}$ act as a wide-baseline stereo system, and thereby have more occlusions but provide accurate depth at far range. The network outputs occlusion mask $M$, small-baseline disparity $d_s$, and disparity $d_m$. The purpose of occlusion mask is to find the occluded pixels in the disparity $d_m$. The disparities at these occluded pixels are replaced by the corresponding small-baseline disparities using disparity assignment loss. This is based on the assumption that occlusion is not a serious problem for small-baseline stereo systems. In the remaining text, $d_m$ will indicate multi-baseline disparity.

\subsection{Proposed Network}
\begin{figure*}[t!]
\scriptsize
    \centering
    \includegraphics[width=\textwidth]{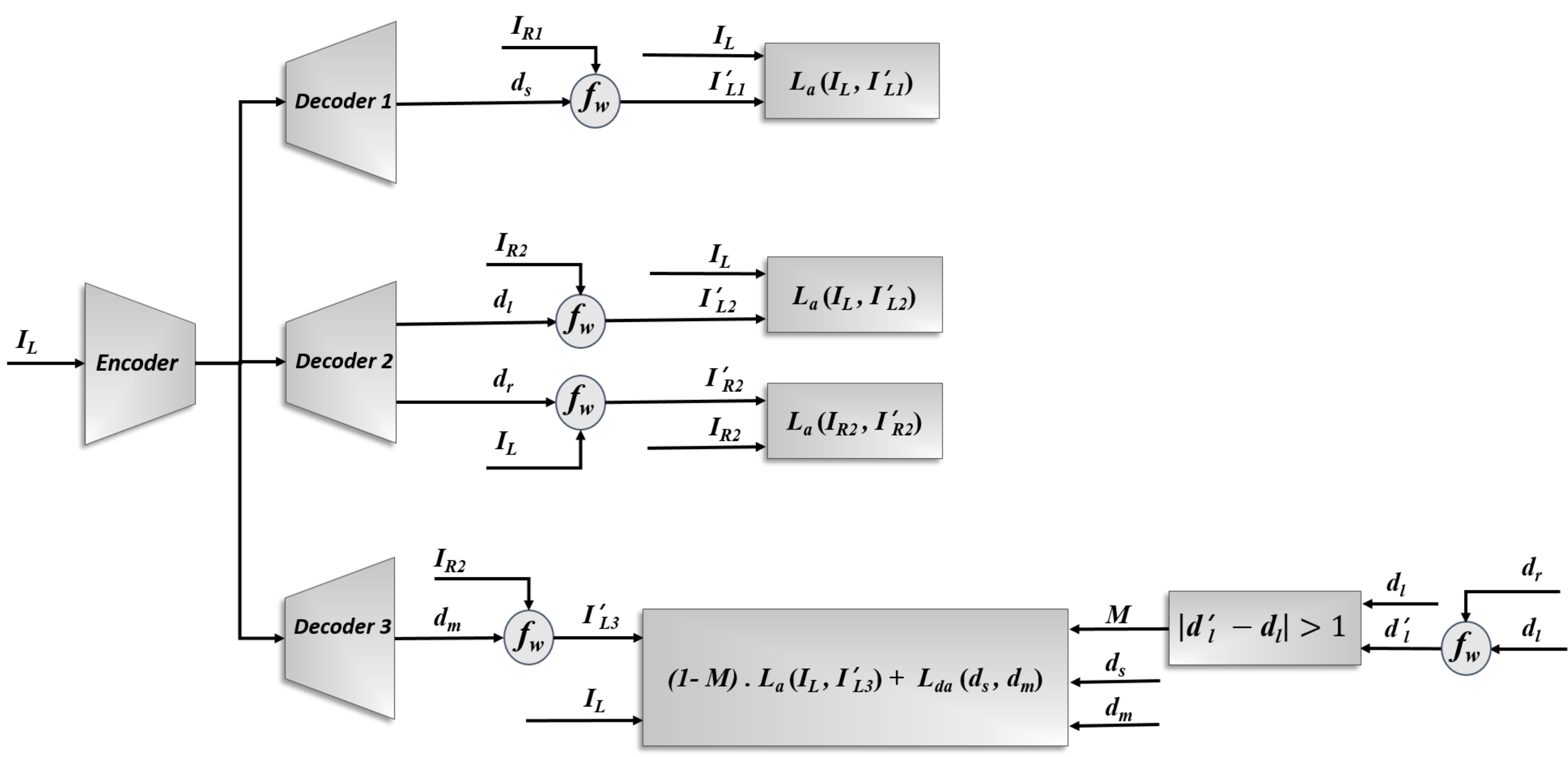}
    \caption{Illustration of our multi-baseline approach. We use three decoders for training. At test time, we only use Decoder 3 to output disparity $d_{m}$. Occlusion mask $M$ is computed using left-right consistency check. The function $f_w$ performs warping operation. }
    \label{fig:network}
\end{figure*}

Given the three rectified images during training, our network learns to infer depth from a single image at test time. As depicted in Figure \ref{fig:network}, the left image $I_{L}$ is fed into the network. We use a shared encoder and three decoders to train the model. Each of the decoders has its own purpose. Decoder 1 utilizes small-baseline image pair $I_{L}$ and $I_{R1}$ to generate small-baseline disparity map $d_{s}$, whereas Decoder 2 uses wide-baseline stereo images $I_{L}$ and $I_{R2}$ for self-supervision to generate left disparity $d_{l}$ and right disparity $d_{r}$, respectively. Multi-baseline disparity $d_{m}$ is generated by Decoder 3, which makes use of the image $I_{L}$, small-baseline disparity $d_{s}$, and the occlusion mask $M$ for supervision.

Similar to \cite{guo2018learning}, we compute the occlusion mask $M$ using the left-right consistency check between the output disparities of Decoder 2,
\begin{equation}
    M=|d'_{l} - d_{l}|>1,
\end{equation}
\noindent where $d'_{l}$ is obtained by warping $d_{r}$ to $d_{l}$. In occluded areas, disparities will have different values \cite{georgoulas2009real}; therefore, we set the threshold to greater than 1 pixel. We assign 1 and 0 to occluded and non-occluded pixels, respectively.

Our network architecture is similar to the encoder-decoder architecture of \cite{godard2017unsupervised}. The encoder is based on VGG \cite{simonyan2014very} and all the decoders have same architecture, except Decoder 2 has two output channels for the disparity maps. Each decoder outputs disparity maps at four scales: one-eighth, quarter, half and full resolutions. Note that the three decoders are only needed for training. Only Decoder 3 is used at inference.
\begin{table*}[t]
\scriptsize
\renewcommand{\arraystretch}{1}

\begin{center}
\begin{tabular}{|c|c|c|c|c|c|c|c|c|}
\hline
 && \multicolumn{4}{|c|}{Lower the better} & \multicolumn{3}{|c|}{Higher the better} \\
\cline{3-6} \cline{7-9}
Method & Baseline & $AbsRel$ & $SqRel$ & $RMSE$ &  $RMSELog$ & $\delta < 1.25$ & $\delta < 1.25^{2}$ & $\delta < 1.25^{3}$\\
\hline
\multicolumn{9}{|c|}{Depth Range = 0.1-10 $m$} \\

  \hline
  Monodepth \cite{godard2017unsupervised}+pp & 10$cm$ &  0.2378&    14.9776&      9.321&      0.311&            0.962&      0.973&      0.976 \\
    Monodepth\cite{godard2017unsupervised}+pp & 54$cm$ &     0.3667&    24.9676&     10.709&      0.387&           0.953&      0.966&      0.971\\
    3Net \cite{poggi2018learning}+pp & 10$cm$ & 0.2076&    13.3266&      8.868&      0.298&           0.970&      0.977&      0.979\\
  3Net \cite{poggi2018learning}+pp & 54$cm$ &  0.9180&    65.9368&     14.974&      0.612&            0.920&      0.935&      0.942\\
  monoResMatch \cite{tosi2019learning}+pp & 10$cm$ &0.5110&    35.2223&      9.324&      0.330&           0.958&      0.966&      0.969\\
  monoResMatch \cite{tosi2019learning}+pp & 54$cm$ & 0.1784&    11.8352&      7.499&      0.250&            0.974&      0.981&      0.984\\
  Monodepth2 \cite{godard2019digging}&10$cm$&0.1182&     7.0760&      6.892&      0.221&      0.980&      0.986&      0.988 \\ 
 Monodepth2 \cite{godard2019digging}&54$cm$&0.1863&    12.5571&      6.949&      0.238&           0.975&      0.982&      0.984 \\ 
  Ours &10$cm$,54$cm$ &  0.1232&     7.6787&      6.273&      0.207&           0.979&      0.986&      0.988\\

\hline
\multicolumn{9}{|c|}{Depth Range = 10-80 $m$} \\

  \hline
  Monodepth \cite{godard2017unsupervised}+pp & 10$cm$ &  0.2027&     3.8602&      9.929&      0.311&          0.727&      0.883&      0.940 \\
    Monodepth\cite{godard2017unsupervised}+pp & 54$cm$ &     0.1772&     3.0741&      9.834&      0.321&           0.753&      0.882&      0.934\\
    3Net \cite{poggi2018learning}+pp & 10$cm$ & 0.2105&     5.2326&     10.102&      0.301&          0.771&      0.898&      0.942\\
  3Net \cite{poggi2018learning}+pp & 54$cm$ &   0.1391&     2.3609&      8.347&      0.273&           0.829&      0.918&      0.953\\
  monoResMatch \cite{tosi2019learning}+pp & 10$cm$ & 0.3239&    10.6855&     13.157&      0.377&           0.667&      0.843&      0.908\\
  monoResMatch \cite{tosi2019learning}+pp & 54$cm$ &0.1677&     4.7352&      9.406&      0.287&           0.838&      0.912&      0.945\\
  Monodepth2 \cite{godard2019digging}&10$cm$& 0.1932&     4.7333&      9.930&      0.284&         0.784&      0.906&      0.947\\
   Monodepth2 \cite{godard2019digging}&54$cm$& 0.1223&     2.3100&      7.896&      0.240&        0.864&      0.937&      0.963 \\ 
  Ours &10$cm$,54$cm$ &  0.1276&     2.0940&      7.967&      0.255&           0.843&      0.928&      0.960\\

\hline
\multicolumn{9}{|c|}{Depth Range = 0.1-80 $m$} \\
  \hline\
 Monodepth\cite{godard2017unsupervised} & 10$cm$ & 0.1520&     4.4573&      7.028&      0.259&          0.890&      0.949&      0.969\\
  Monodepth \cite{godard2017unsupervised}+pp & 10$cm$ & 0.1023&     1.5862&      5.804&      0.202&           0.891&      0.954&      0.976 \\
  Monodepth \cite{godard2017unsupervised} & 54$cm$ & 0.3192&    12.3581&      8.279&      0.397&           0.870&      0.921&      0.944\\
    Monodepth\cite{godard2017unsupervised}+pp & 54$cm$ &    0.1364&     3.5626&      6.150&      0.250&          0.892&      0.947&      0.969\\
    3Net \cite{poggi2018learning} & 10$cm$ & 0.1047&     2.2487&      6.036&      0.203&          0.909&      0.959&      0.977\\
    3Net \cite{poggi2018learning}+pp & 10$cm$ & 0.1028&     2.1953&      5.950&      0.201&          0.911&      0.960&      0.977\\
     3Net \cite{poggi2018learning} & 54$cm$ & 0.2794&    11.4045&      7.255&      0.361&         0.897&      0.938&      0.956\\
  3Net \cite{poggi2018learning}+pp & 54$cm$ &  0.2779&    11.3585&      7.214&      0.360&          0.898&      0.938&      0.956\\
  
   monoResMatch \cite{tosi2019learning} & 10$cm$ &0.4658&    25.1243&     10.801&      0.377&          0.824&      0.904&      0.939\\
  monoResMatch \cite{tosi2019learning}+pp & 10$cm$ &0.3711&    20.1240&      8.948&      0.299&          0.864&      0.930&      0.954\\
  monoResMatch \cite{tosi2019learning} & 54$cm$ &0.3129&    12.9569&      8.748&      0.390&           0.884&      0.925&      0.945\\
  monoResMatch \cite{tosi2019learning}+pp & 54$cm$ &0.1004&     3.3953&      5.770&      0.194&             0.933&      0.964&      0.977\\
  Monodepth2 \cite{godard2019digging}&10$cm$& 0.0843&     1.7101&      5.781&      0.184&         0.918&      0.964&      0.979\\
   Monodepth2 \cite{godard2019digging}&54$cm$&0.0864&     2.3483&      4.969&      0.186&         \bfseries0.939&      0.969&      0.981  \\ 
  Ours &10$cm$,54 $cm$ & \bfseries0.0643&    \bfseries 0.9509&      \bfseries4.695&     \bfseries 0.163&          0.936&     \bfseries 0.971&      \bfseries0.984 \\

  \hline
\end{tabular}
\end{center}
\caption{ Evaluation on CARLA dataset. For comparison, we train previous methods separately with 10 $cm$ and 54 $cm$ baseline stereo images. Maximum predictions of all the networks are capped to 80 $m$. pp stands for post-processing.}
\label{tab:comp_wopp}
\end{table*}
\subsection{Training Losses}
We train our network with multiple losses. In addition to image reconstruction loss $L_{a}$ and disparity smoothness loss $L_{s}$, we also use disparity assignment loss $L_{da}$. All the losses are minimized at four scales. 

\textbf{Image Reconstruction Loss} Following \cite{godard2017unsupervised}, we use the weighted sum of SSIM \cite{wang2004image} and L1 loss to minimize the photometric error between the reconstructed and original images,

\begin{multline}
    L_{a}(I,I')=\frac{1}{N} \sum_{i, j}\bigg(\alpha \frac{1 - SSIM(I_{ij},I'_{ij})}{2} \\+ (1-\alpha)|I_{ij} - I'_{ij}|\bigg),
\end{multline}

\noindent where $N$ is the number of pixels, and $I$ and $I'$ are the original and reconstructed images. $\alpha$ is set to 0.85. 

\textbf{Disparity Smoothness Loss} Similar to \cite{godard2017unsupervised}, we define an edge-aware smoothness loss to deal with disparity discontinuities.
\begin{eqnarray}
  L_{s}(d,I)=\frac{1}{N}\sum_{i,j}|\partial_{x} d_{ij}|e^{-|\partial_{x} I_{ij}|}+|\partial_{y}d_{ij}|e^{-|\partial_{y} I_{ij}|}.
\end{eqnarray}
Here $d$ is the disparity, $I$ is the corresponding image, and $\partial_{x}$, $\partial_{y}$ are the horizontal and vertical gradients, respectively. 

\textbf{Disparity Assignment Loss} To replace the occluded pixels of the  disparity map $d_{m}$ (Fig. \ref{fig:network}) by the small-baseline disparity map $d_{s}$, we again employ the combination of L1 and SSIM losses as

\begin{multline}
     L_{da}(d_{s},d_{m})=M \cdot \frac{1}{N} \sum_{i, j}\bigg(\beta \frac{1 -SSIM(r \cdot d_{s},d_{m})}{2} \\
      + (1-\beta)|r\cdot d_{s}-d_{m}| \bigg),
\end{multline}
where $\beta$ is the weighting factor set to 0.85, and $r$ is the ratio of wide baseline to small baseline. The occlusion mask $M$ ensures that only occluded pixels are considered. The factor $r$ is used to scale the disparity $d_{s}$ to match the disparity range of $d_{m}$. To ensure that the disparity $d_{m}$ follows the disparity $d_{s}$ in occluded regions, the gradients for this loss function are not computed with respect to $d_{s}$. In other words, only the weights of Decoder 3 will change to minimize $L_{da}$. The total loss is given as follows
\begin{multline}
    L_{total}=L_{a}(I_{L},I'_{L1})  + L_{a}(I_{L},I'_{L2}) + L_{a}(I_{R2},I'_{R2}) + \\ \lambda ( L_{s}(d_{s},I_{L}) + L_{s}(d_{l},I_{L}) +  L_{s}(d_{r},I_{R2}) ) + L_{dec3},
\end{multline}

\noindent where $\lambda$ is the weighting factor set to 0.1, and the Decoder 3 loss $L_{dec3}$ is defined as
\begin{equation}\label{dec3}
    L_{dec3}=(1-M) \cdot L_{a}(I_{L},I'_{L3})+L_{da}(d_{s},d_{m})+\lambda \cdot L_{s}(d_{m},I_{L}).
\end{equation}
In Eq. \ref{dec3}, the term $(1-M)$ ensures that the occluded pixels do not contribute to the image reconstruction loss $L_{a}(I_{L},I'_{L3})$. The occluded pixels are filled using $L_{da}$ loss only.
\begin{figure*}[t!]
\tiny
     \centering
     \setlength{\tabcolsep}{0mm}
    \begin{tabular}{cccccccc}
 
        &
           	\begin{tabular}{@{}c}\rotatebox{90}{Input}\end{tabular}
		  & 
            	\begin{tabular}{@{}c}\includegraphics[width=0.2\linewidth,height=0.1\linewidth,keepaspectratio,]{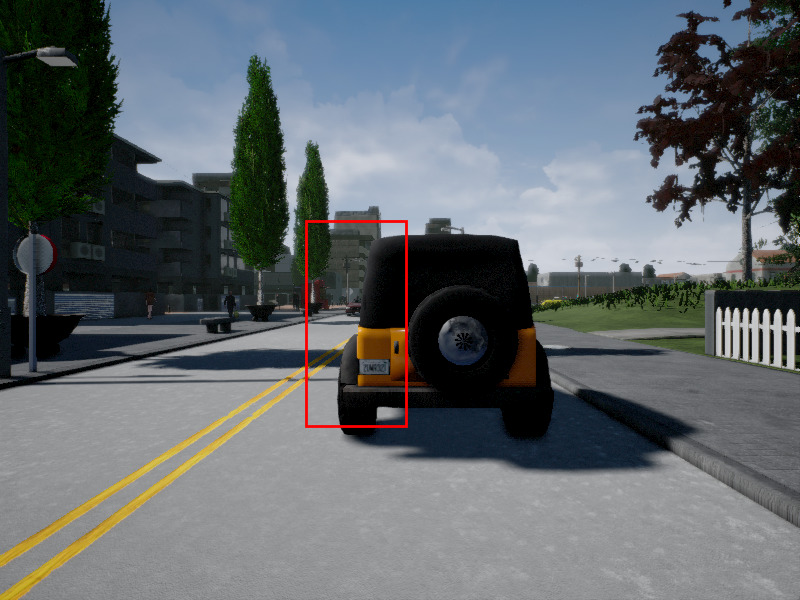}\end{tabular}\hspace{0mm}\hfill
		  &
		  \begin{tabular}{@{}c}\includegraphics[width=0.125\linewidth,height=0.1\linewidth,keepaspectratio,]{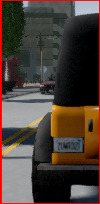}\end{tabular}\hspace{1.5mm}
		  & 
           	\begin{tabular}{@{}c}\includegraphics[width=0.2\linewidth,height=0.1\linewidth,keepaspectratio,]{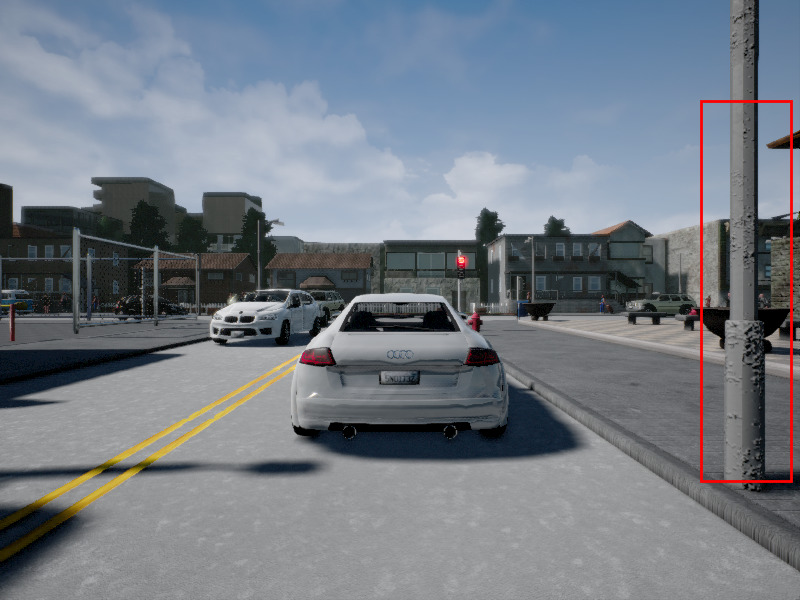}\end{tabular}
		  & 
		  \begin{tabular}{@{}c}\includegraphics[width=0.125\linewidth,height=0.1\linewidth,keepaspectratio,]{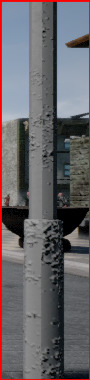}\end{tabular}\hspace{1.5mm}
		  & 
    		\begin{tabular}{@{}c}\includegraphics[width=0.2\linewidth,height=0.1\linewidth,keepaspectratio,]{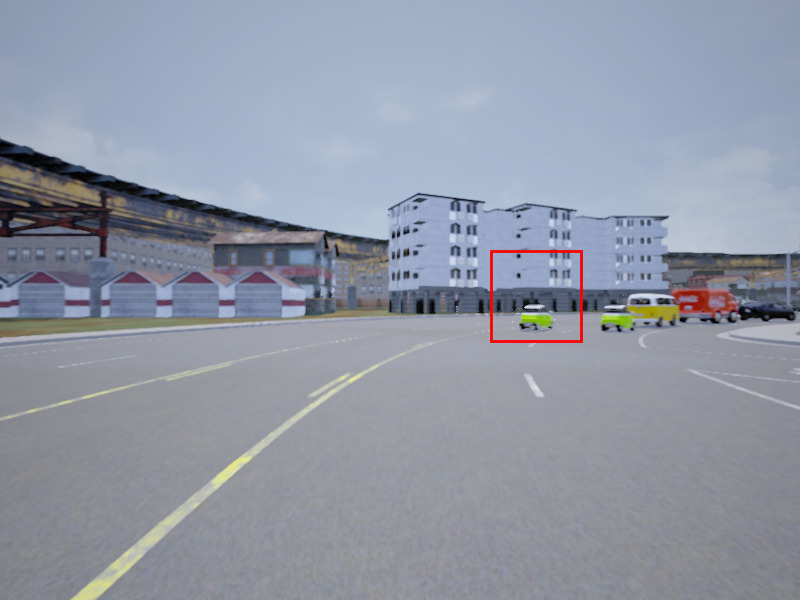}\end{tabular}
		  & 
		  \begin{tabular}{@{}c}\includegraphics[width=0.125\linewidth,height=0.1\linewidth,keepaspectratio,]{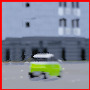}\end{tabular}\hspace{0.75mm}
		  \\
		  
		  		  &
		             	\begin{tabular}{@{}c}\rotatebox{90}{Ground Truth}\end{tabular}
		  & 
            	\begin{tabular}{@{}c}\includegraphics[width=0.2\linewidth,height=0.1\linewidth,keepaspectratio,]{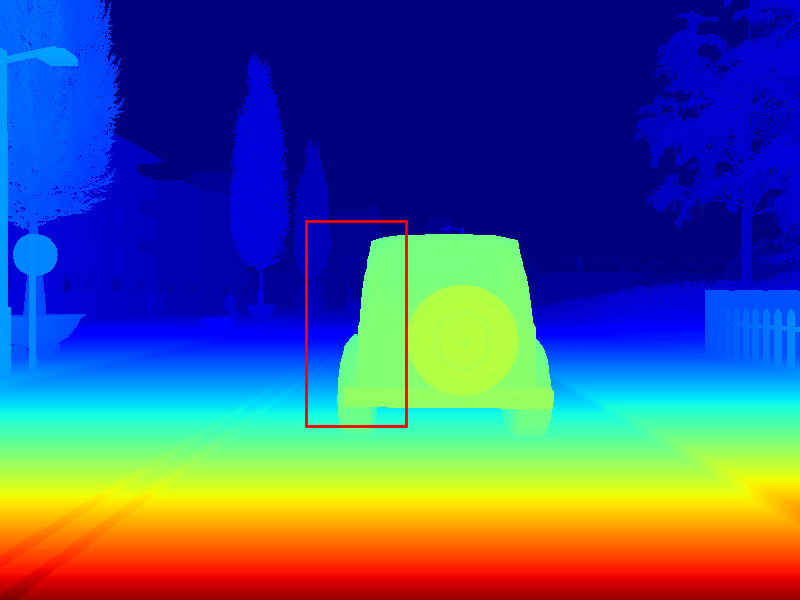}\end{tabular}\hspace{0mm}\hfill
		  &
		  \begin{tabular}{@{}c}\includegraphics[width=0.125\linewidth,height=0.1\linewidth,keepaspectratio,]{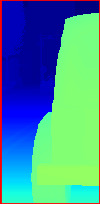}\end{tabular}\hspace{1.5mm}
		  & 
          	\begin{tabular}{@{}c}\includegraphics[width=0.2\linewidth,height=0.1\linewidth,keepaspectratio,]{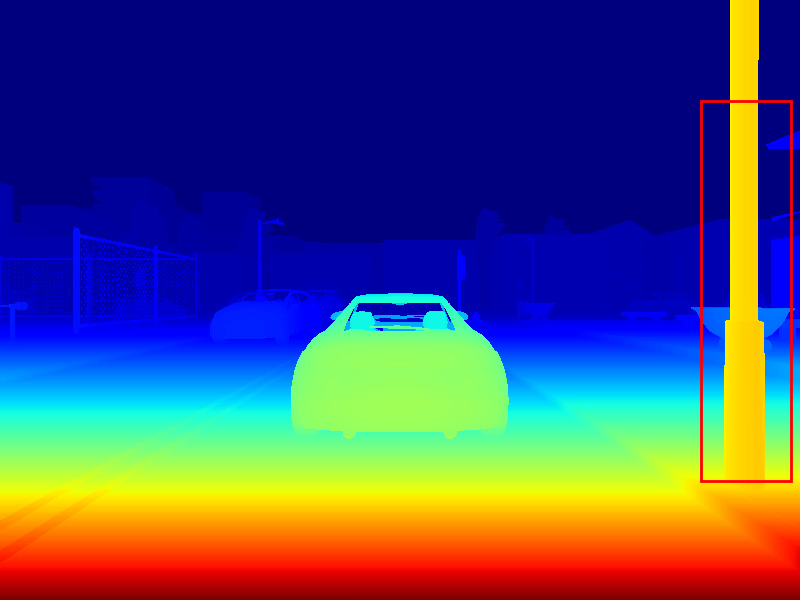}\end{tabular}
		  & 
		  \begin{tabular}{@{}c}\includegraphics[width=0.125\linewidth,height=0.1\linewidth,keepaspectratio,]{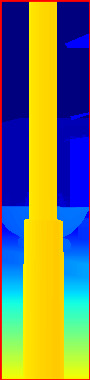}\end{tabular}\hspace{1.5mm}
		  & 
    		\begin{tabular}{@{}c}\includegraphics[width=0.2\linewidth,height=0.1\linewidth,keepaspectratio,]{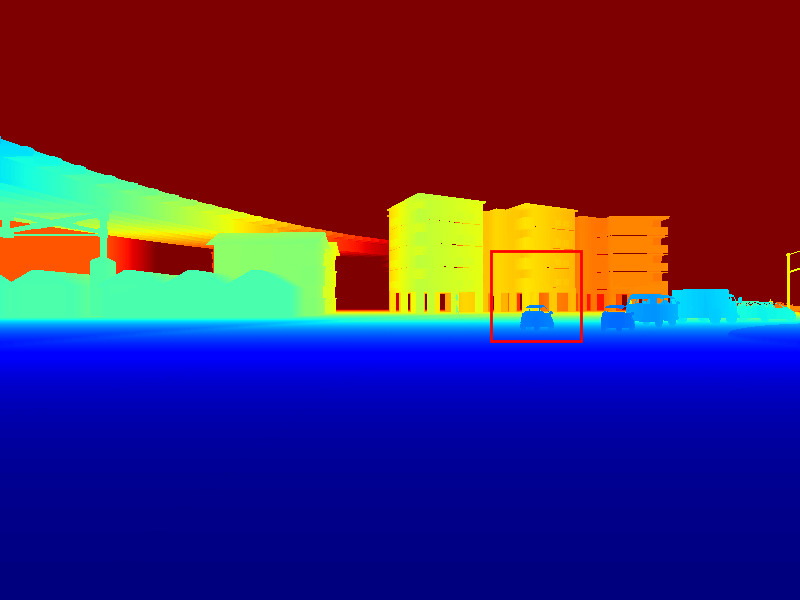}\end{tabular}
		  & 
		  \begin{tabular}{@{}c}\includegraphics[width=0.125\linewidth,height=0.1\linewidth,keepaspectratio,]{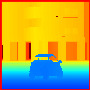}\end{tabular}\hspace{0.75mm}
		  \\

		             	\begin{tabular}{@{}c}\rotatebox{90}{3net}\end{tabular}
		  & 
		  	\begin{tabular}{@{}c}\rotatebox{90}{(54cm) \cite{poggi2018learning}}\end{tabular}
		  	&
            	\begin{tabular}{@{}c}\includegraphics[width=0.2\linewidth,height=0.1\linewidth,keepaspectratio,]{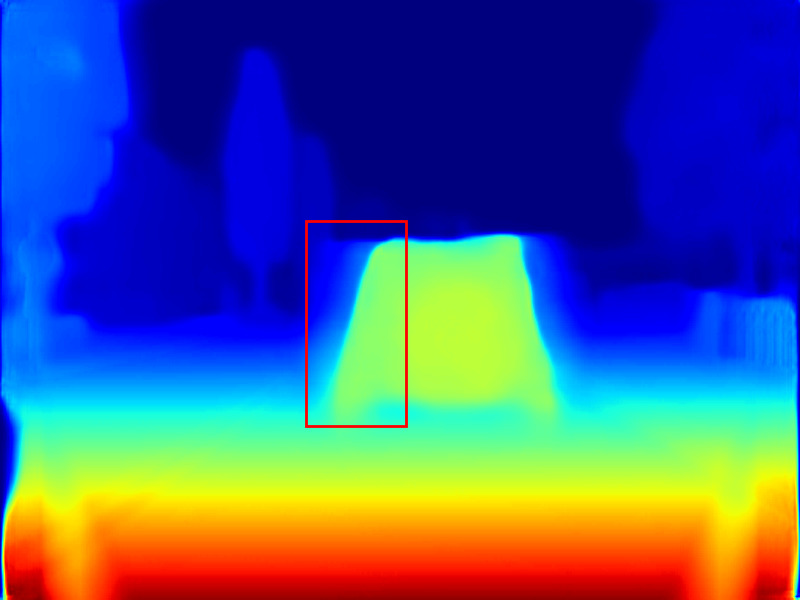}\end{tabular}\hspace{0mm}\hfill
		  &
		  \begin{tabular}{@{}c}\includegraphics[width=0.125\linewidth,height=0.1\linewidth,keepaspectratio,]{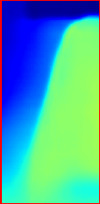}\end{tabular}\hspace{1.5mm}
		  & 
           	\begin{tabular}{@{}c}\includegraphics[width=0.2\linewidth,height=0.1\linewidth,keepaspectratio,]{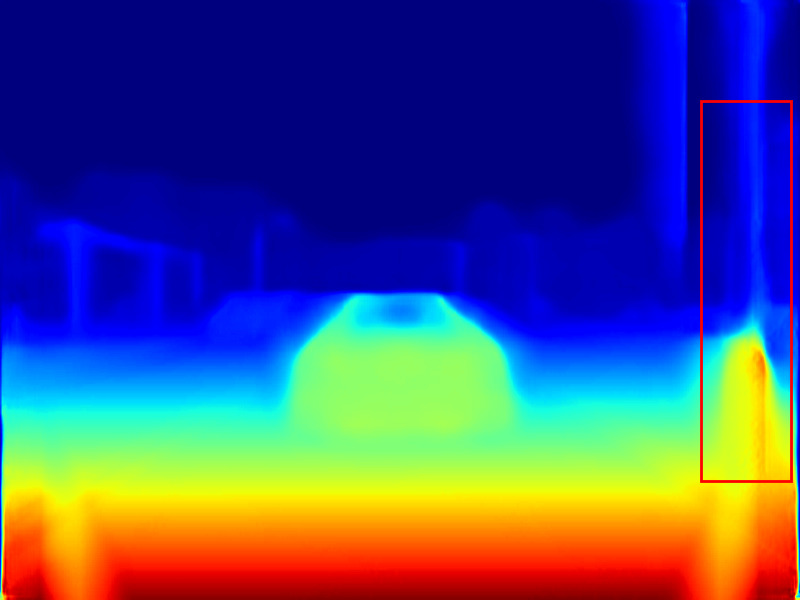}\end{tabular}
		  & 
		  \begin{tabular}{@{}c}\includegraphics[width=0.125\linewidth,height=0.1\linewidth,keepaspectratio,]{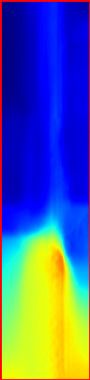}\end{tabular}\hspace{1.5mm}
		  & 
    		\begin{tabular}{@{}c}\includegraphics[width=0.2\linewidth,height=0.1\linewidth,keepaspectratio,]{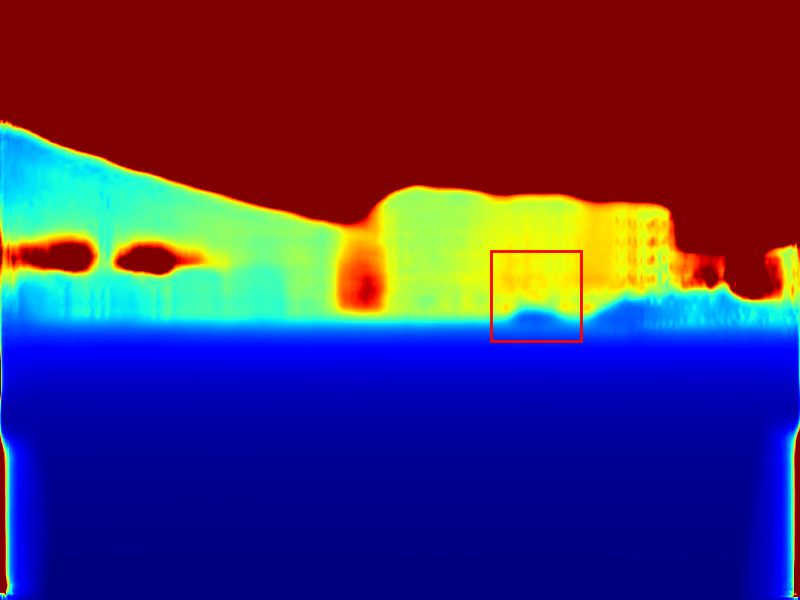}\end{tabular}
		  & 
		  \begin{tabular}{@{}c}\includegraphics[width=0.125\linewidth,height=0.1\linewidth,keepaspectratio,]{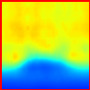}\end{tabular}\hspace{0.75mm}
		  \\
		  
		             	\begin{tabular}{@{}c}\rotatebox{90}{3net}\end{tabular}
		  & 
		  	\begin{tabular}{@{}c}\rotatebox{90}{(10cm) \cite{poggi2018learning}}\end{tabular}
		  & 
            	\begin{tabular}{@{}c}\includegraphics[width=0.2\linewidth,height=0.1\linewidth,keepaspectratio,]{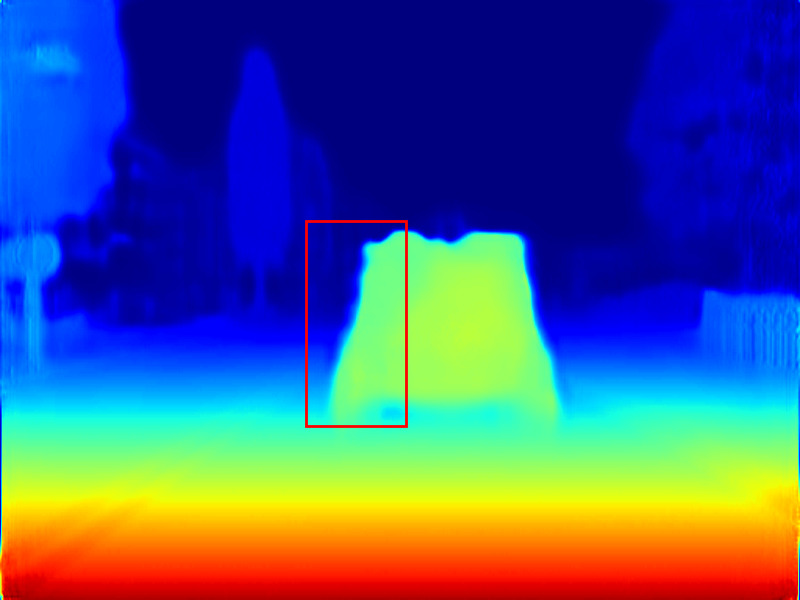}\end{tabular}\hspace{0mm}\hfill
		  &
		  \begin{tabular}{@{}c}\includegraphics[width=0.125\linewidth,height=0.1\linewidth,keepaspectratio,]{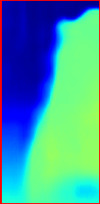}\end{tabular}\hspace{1.5mm}
		  & 
          	\begin{tabular}{@{}c}\includegraphics[width=0.2\linewidth,height=0.1\linewidth,keepaspectratio,]{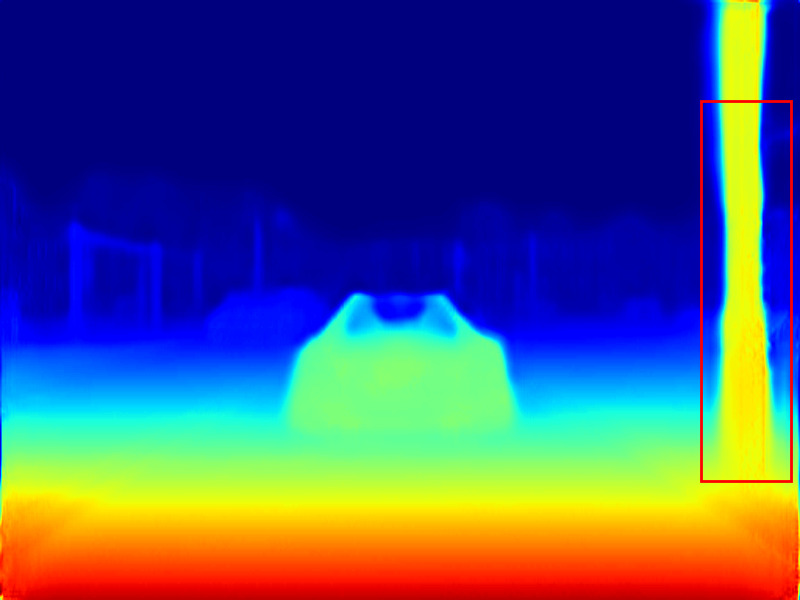}\end{tabular}
		  & 
		  \begin{tabular}{@{}c}\includegraphics[width=0.125\linewidth,height=0.1\linewidth,keepaspectratio,]{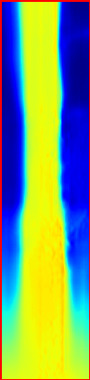}\end{tabular}\hspace{1.5mm}
		  & 
    		\begin{tabular}{@{}c}\includegraphics[width=0.2\linewidth,height=0.1\linewidth,keepaspectratio,]{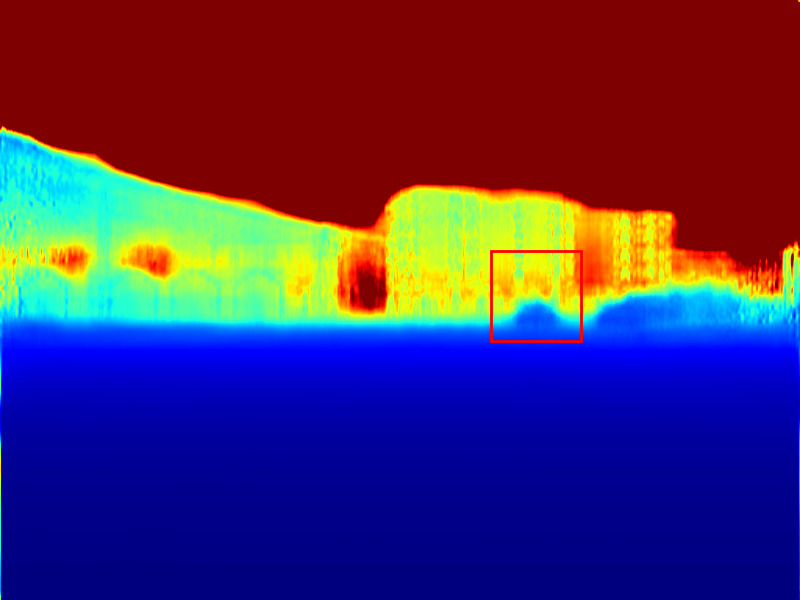}\end{tabular}
		  & 
		  \begin{tabular}{@{}c}\includegraphics[width=0.125\linewidth,height=0.1\linewidth,keepaspectratio,]{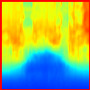}\end{tabular}\hspace{0.75mm}
		  \\
		  
		             	\begin{tabular}{@{}c}\rotatebox{90}{Monodepth}\end{tabular}
		  & 
		  	\begin{tabular}{@{}c}\rotatebox{90}{(54cm) \cite{godard2017unsupervised}}\end{tabular}
		  & 
            	\begin{tabular}{@{}c}\includegraphics[width=0.2\linewidth,height=0.1\linewidth,keepaspectratio,]{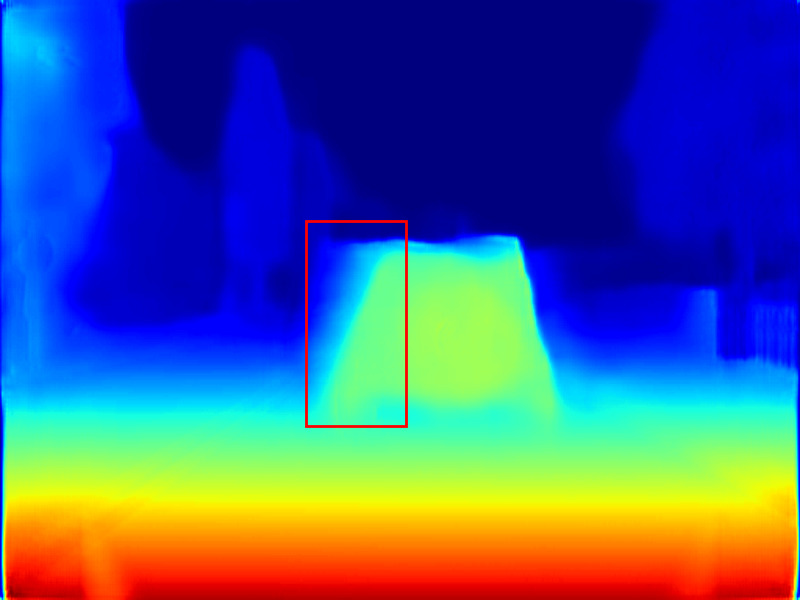}\end{tabular}\hspace{0mm}\hfill
		  &
		  \begin{tabular}{@{}c}\includegraphics[width=0.125\linewidth,height=0.1\linewidth,keepaspectratio,]{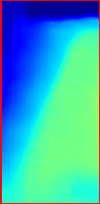}\end{tabular}\hspace{1.5mm}
		  & 
          	\begin{tabular}{@{}c}\includegraphics[width=0.2\linewidth,height=0.1\linewidth,keepaspectratio,]{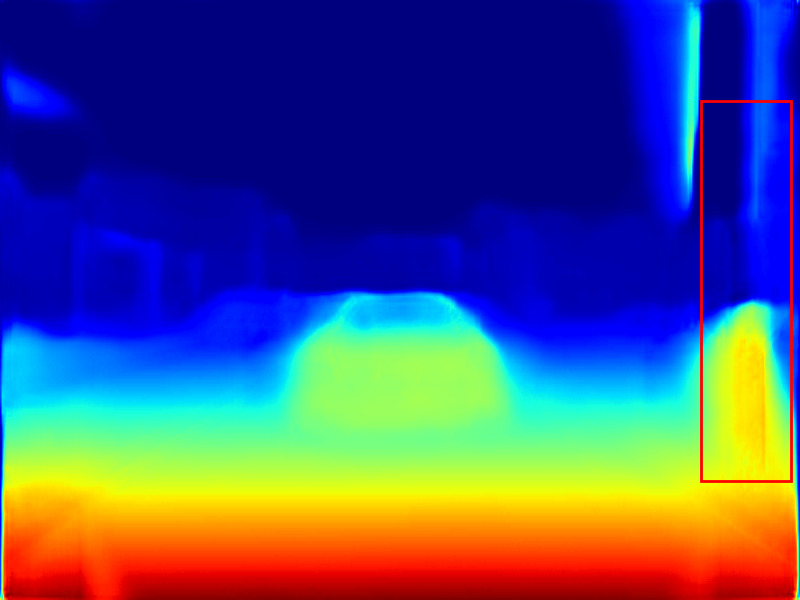}\end{tabular}
		  & 
		  \begin{tabular}{@{}c}\includegraphics[width=0.125\linewidth,height=0.1\linewidth,keepaspectratio,]{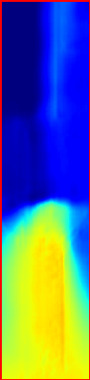}\end{tabular}\hspace{1.5mm}
		  & 
    		\begin{tabular}{@{}c}\includegraphics[width=0.2\linewidth,height=0.1\linewidth,keepaspectratio,]{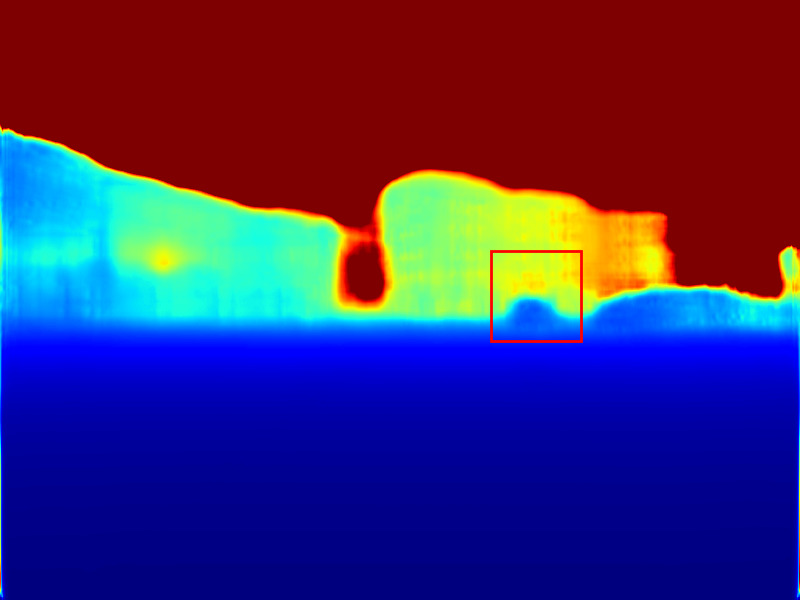}\end{tabular}
		  & 
		  \begin{tabular}{@{}c}\includegraphics[width=0.125\linewidth,height=0.1\linewidth,keepaspectratio,]{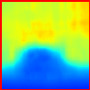}\end{tabular}\hspace{0.75mm}
		  \\
		  
		             	\begin{tabular}{@{}c}\rotatebox{90}{Monodepth}\end{tabular}
		  & 
		  \begin{tabular}{@{}c}\rotatebox{90}{(10cm) \cite{godard2017unsupervised}}\end{tabular}
		  & 
            	\begin{tabular}{@{}c}\includegraphics[width=0.2\linewidth,height=0.1\linewidth,keepaspectratio,]{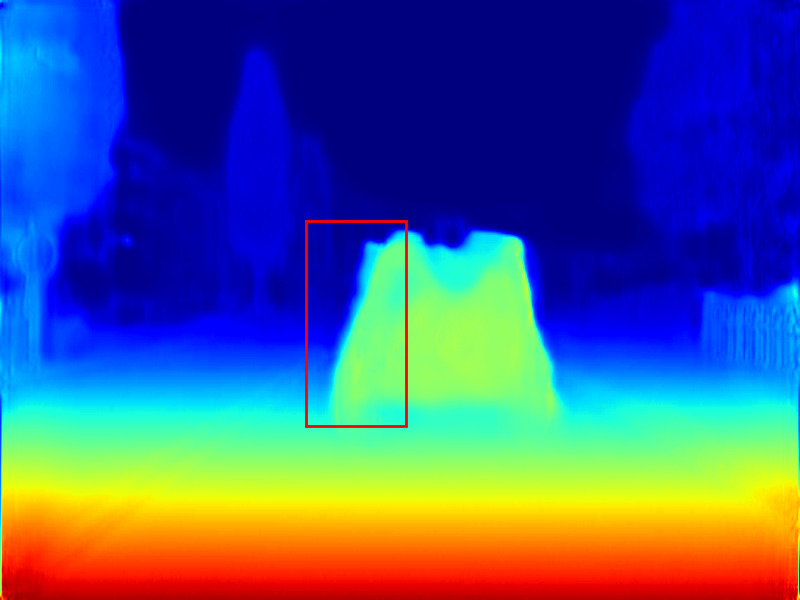}\end{tabular}\hspace{0mm}\hfill
		  &
		  \begin{tabular}{@{}c}\includegraphics[width=0.125\linewidth,height=0.1\linewidth,keepaspectratio,]{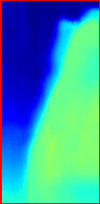}\end{tabular}\hspace{1.5mm}
		  & 
          	\begin{tabular}{@{}c}\includegraphics[width=0.2\linewidth,height=0.1\linewidth,keepaspectratio,]{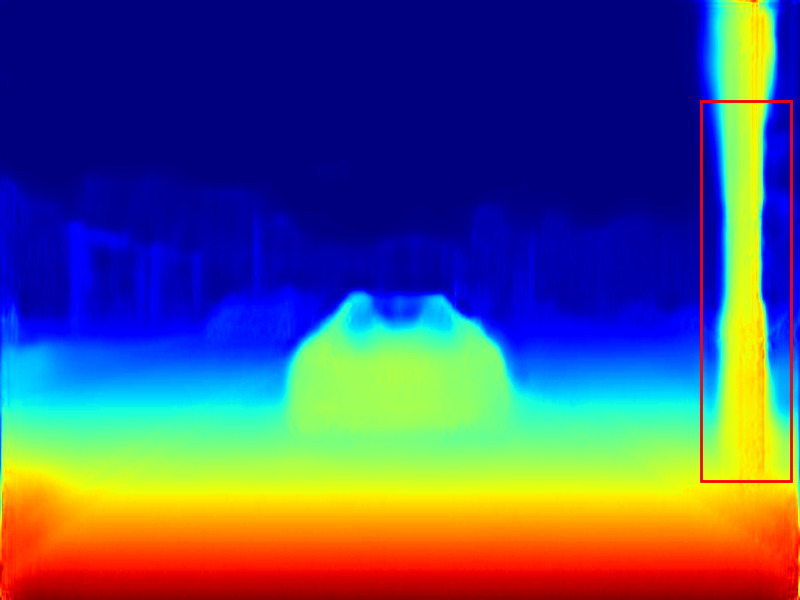}\end{tabular}
		  & 
		  \begin{tabular}{@{}c}\includegraphics[width=0.125\linewidth,height=0.1\linewidth,keepaspectratio,]{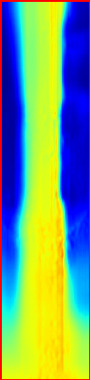}\end{tabular}\hspace{1.5mm}
		  & 
    		\begin{tabular}{@{}c}\includegraphics[width=0.2\linewidth,height=0.1\linewidth,keepaspectratio,]{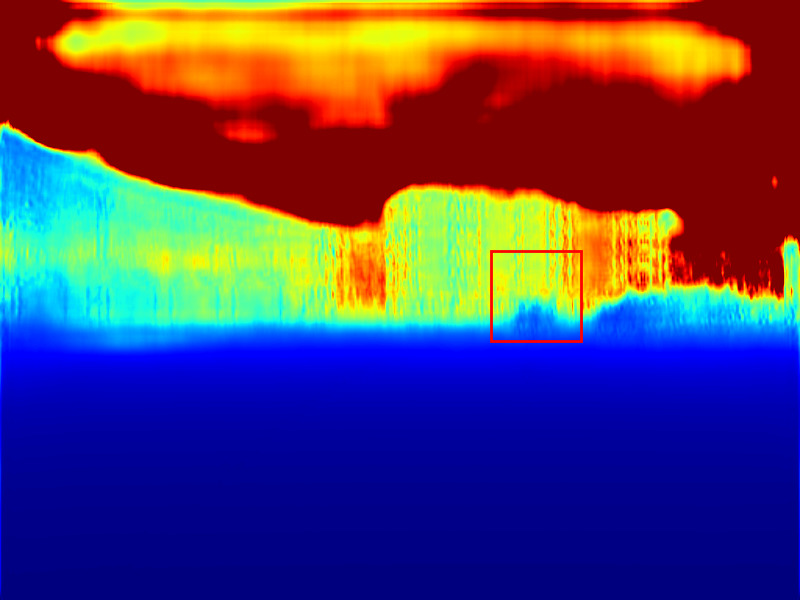}\end{tabular}
		  & 
		  \begin{tabular}{@{}c}\includegraphics[width=0.125\linewidth,height=0.1\linewidth,keepaspectratio,]{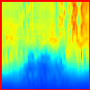}\end{tabular}\hspace{0.75mm}
		  \\

		              	\begin{tabular}{@{}c}\rotatebox{90}{monoResMatch}\end{tabular}
		  & 
		  \begin{tabular}{@{}c}\rotatebox{90}{(54cm) \cite{tosi2019learning}}\end{tabular}
		  & 
            	\begin{tabular}{@{}c}\includegraphics[width=0.2\linewidth,height=0.1\linewidth,keepaspectratio,]{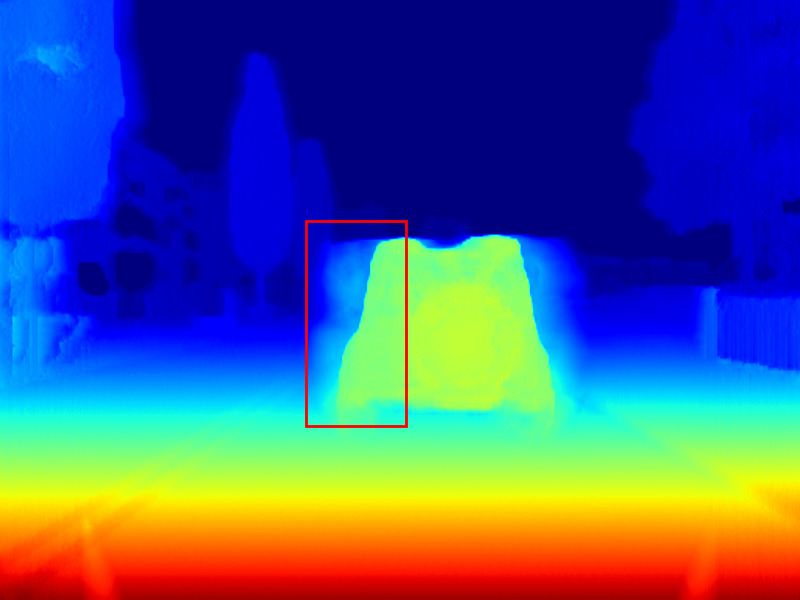}\end{tabular}\hspace{0mm}\hfill
		  &
		  \begin{tabular}{@{}c}\includegraphics[width=0.125\linewidth,height=0.1\linewidth,keepaspectratio,]{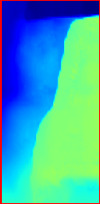}\end{tabular}\hspace{1.5mm}
		  & 
          	\begin{tabular}{@{}c}\includegraphics[width=0.2\linewidth,height=0.1\linewidth,keepaspectratio,]{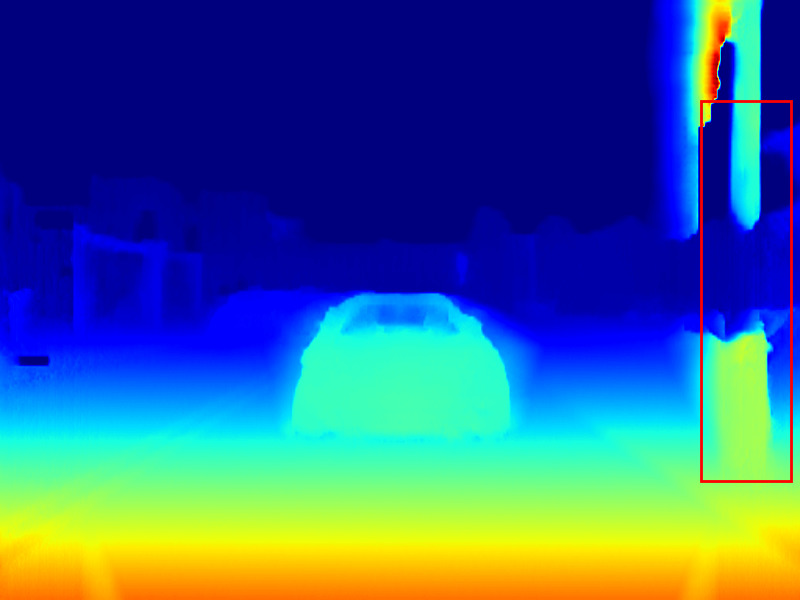}\end{tabular}
		  & 
		  \begin{tabular}{@{}c}\includegraphics[width=0.125\linewidth,height=0.1\linewidth,keepaspectratio,]{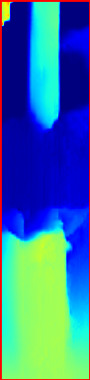}\end{tabular}\hspace{1.5mm}
		  & 
    		\begin{tabular}{@{}c}\includegraphics[width=0.2\linewidth,height=0.1\linewidth,keepaspectratio,]{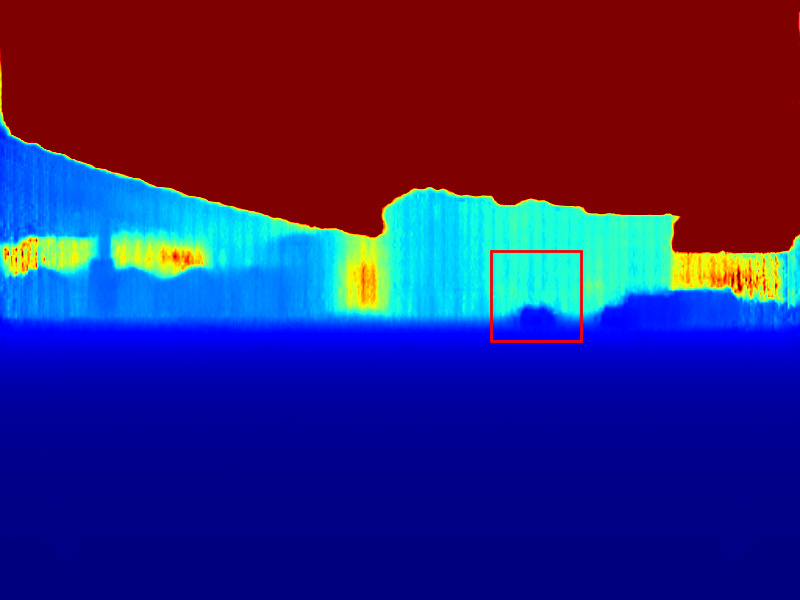}\end{tabular}
		  & 
		  \begin{tabular}{@{}c}\includegraphics[width=0.125\linewidth,height=0.1\linewidth,keepaspectratio,]{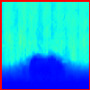}\end{tabular}\hspace{0.75mm}
		  \\
		  
		  \begin{tabular}{@{}c}\rotatebox{90}{monoResMatch}\end{tabular}
		  & 
		           	\begin{tabular}{@{}c}\rotatebox{90}{(10cm) \cite{tosi2019learning}}\end{tabular}
		  & 
            	\begin{tabular}{@{}c}\includegraphics[width=0.2\linewidth,height=0.1\linewidth,keepaspectratio,]{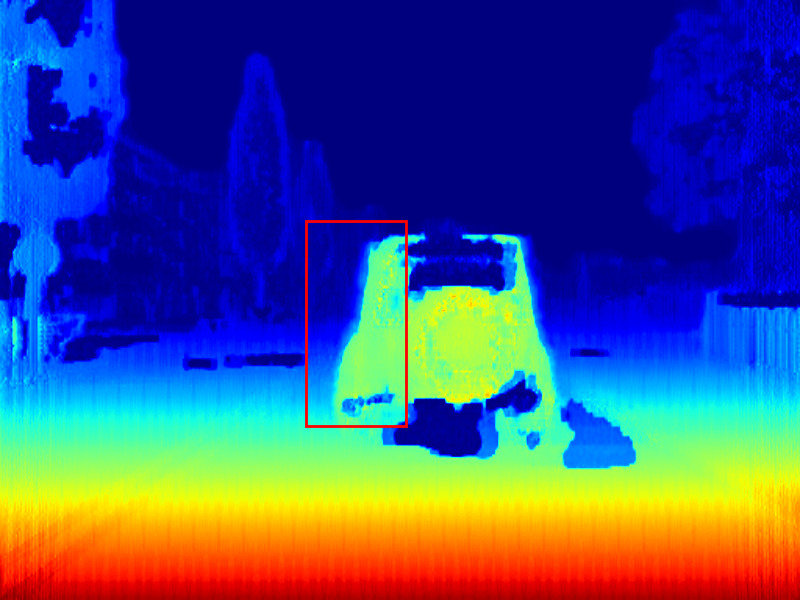}\end{tabular}\hspace{0mm}\hfill
		  &
		  \begin{tabular}{@{}c}\includegraphics[width=0.125\linewidth,height=0.1\linewidth,keepaspectratio,]{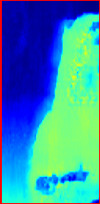}\end{tabular}\hspace{1.5mm}
		  & 
          	\begin{tabular}{@{}c}\includegraphics[width=0.2\linewidth,height=0.1\linewidth,keepaspectratio,]{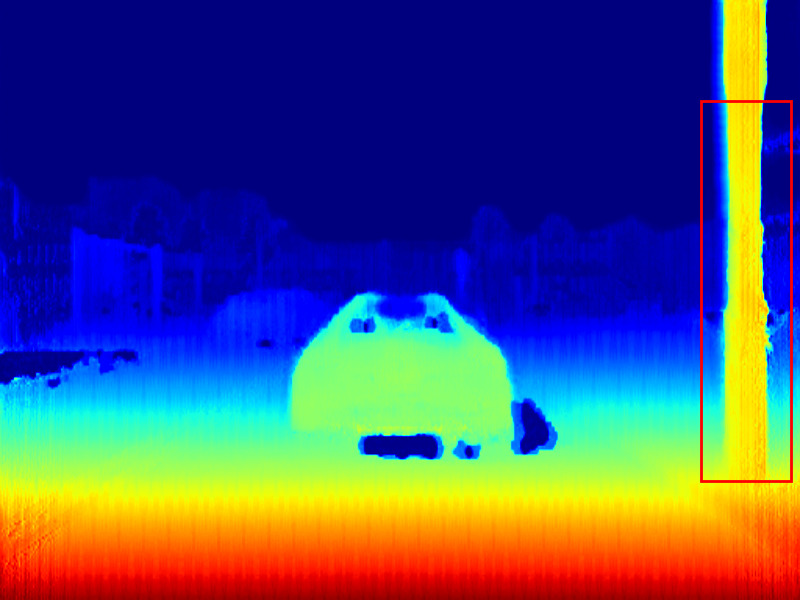}\end{tabular}
		  & 
		  \begin{tabular}{@{}c}\includegraphics[width=0.125\linewidth,height=0.1\linewidth,keepaspectratio,]{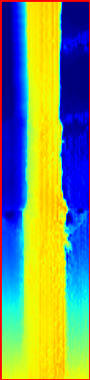}\end{tabular}\hspace{1.5mm}
		  & 
    		\begin{tabular}{@{}c}\includegraphics[width=0.2\linewidth,height=0.1\linewidth,keepaspectratio,]{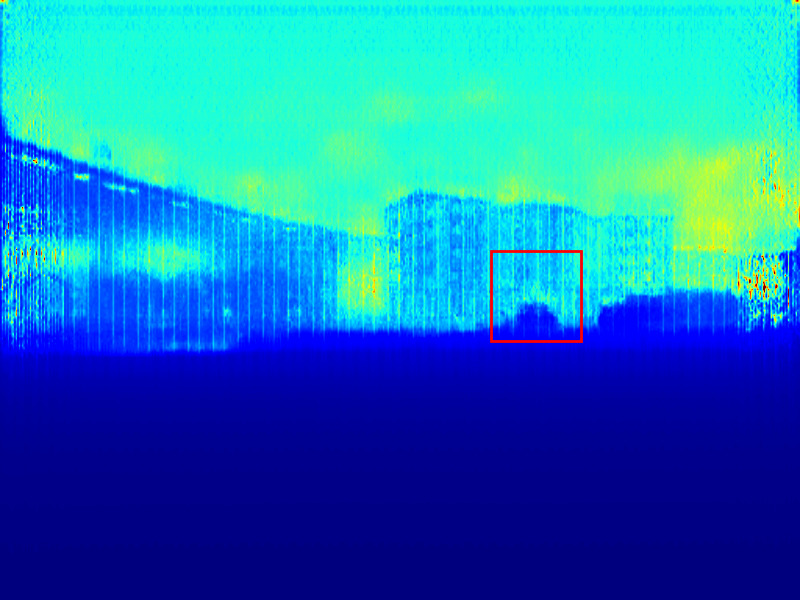}\end{tabular}
		  & 
		  \begin{tabular}{@{}c}\includegraphics[width=0.125\linewidth,height=0.1\linewidth,keepaspectratio,]{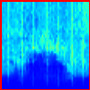}\end{tabular}\hspace{0.75mm}
		  \\
		  		  \begin{tabular}{@{}c}\rotatebox{90}{Monodepth2}\end{tabular}
		  & 
		           	\begin{tabular}{@{}c}\rotatebox{90}{(54cm) \cite{godard2019digging}}\end{tabular}
		  & 
            	\begin{tabular}{@{}c}\includegraphics[width=0.2\linewidth,height=0.1\linewidth,keepaspectratio,]{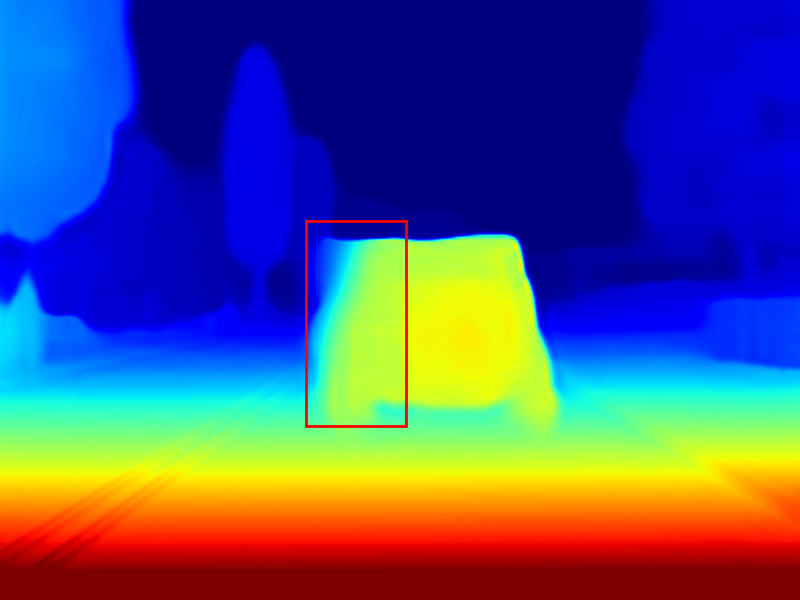}\end{tabular}\hspace{0mm}\hfill
		  &
		  \begin{tabular}{@{}c}\includegraphics[width=0.125\linewidth,height=0.1\linewidth,keepaspectratio,]{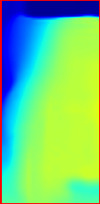}\end{tabular}\hspace{1.5mm}
		  & 
          	\begin{tabular}{@{}c}\includegraphics[width=0.2\linewidth,height=0.1\linewidth,keepaspectratio,]{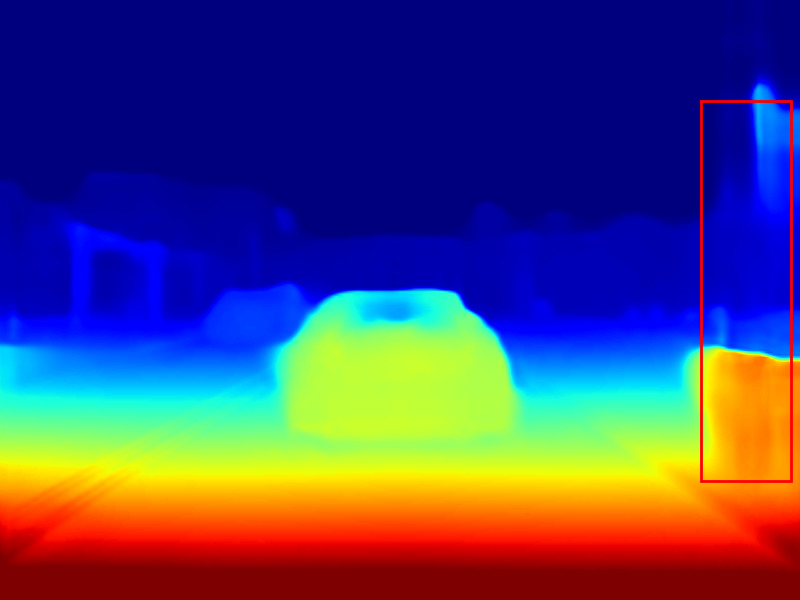}\end{tabular}
		  & 
		  \begin{tabular}{@{}c}\includegraphics[width=0.125\linewidth,height=0.1\linewidth,keepaspectratio,]{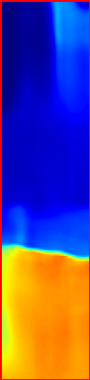}\end{tabular}\hspace{1.5mm}
		  & 
    		\begin{tabular}{@{}c}\includegraphics[width=0.2\linewidth,height=0.1\linewidth,keepaspectratio,]{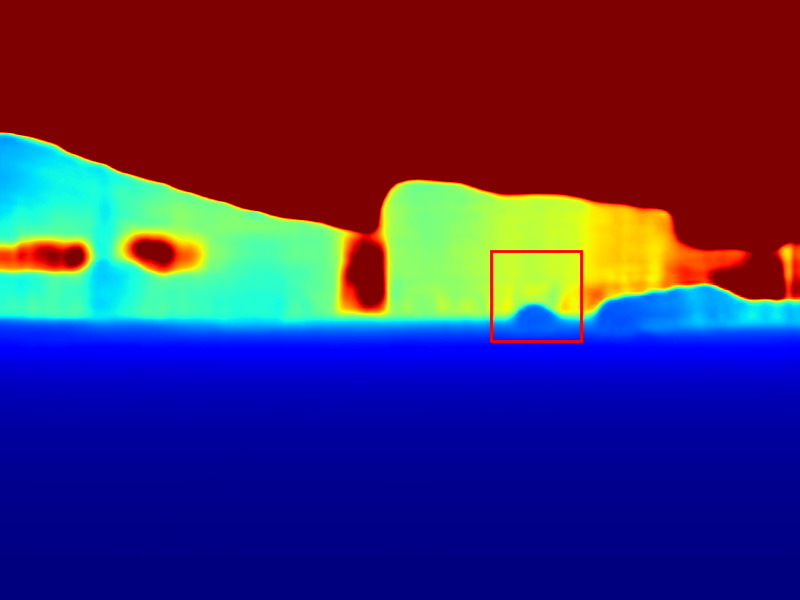}\end{tabular}
		  & 
		  \begin{tabular}{@{}c}\includegraphics[width=0.125\linewidth,height=0.1\linewidth,keepaspectratio,]{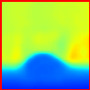}\end{tabular}\hspace{0.75mm}
		  \\
		  	  		  \begin{tabular}{@{}c}\rotatebox{90}{Monodepth2}\end{tabular}
		  & 
		           	\begin{tabular}{@{}c}\rotatebox{90}{(10cm) \cite{godard2019digging}}\end{tabular}
		  & 
            	\begin{tabular}{@{}c}\includegraphics[width=0.2\linewidth,height=0.1\linewidth,keepaspectratio,]{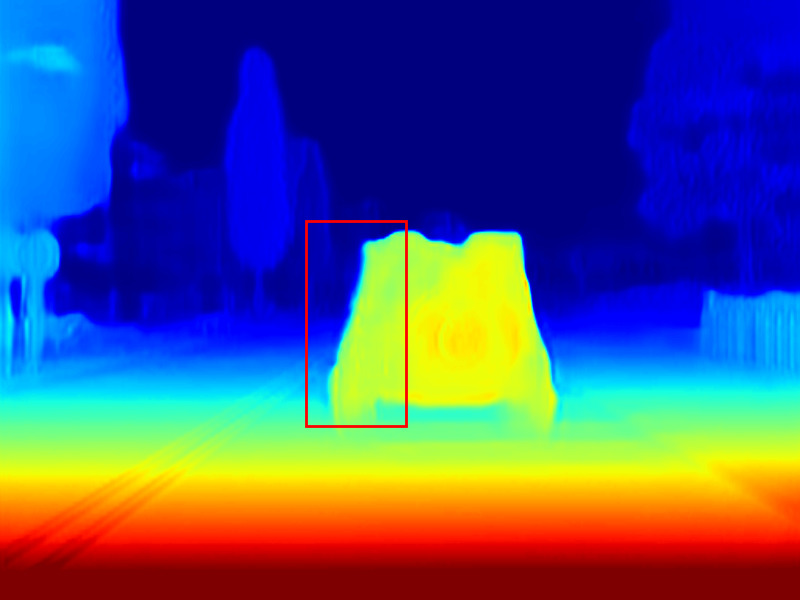}\end{tabular}\hspace{0mm}\hfill
		  &
		  \begin{tabular}{@{}c}\includegraphics[width=0.125\linewidth,height=0.1\linewidth,keepaspectratio,]{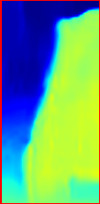}\end{tabular}\hspace{1.5mm}
		  & 
          	\begin{tabular}{@{}c}\includegraphics[width=0.2\linewidth,height=0.1\linewidth,keepaspectratio,]{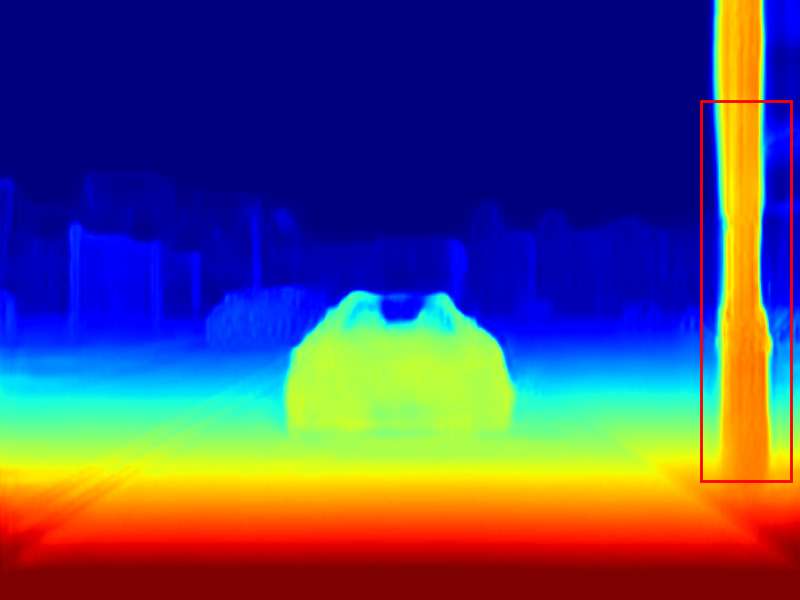}\end{tabular}
		  & 
		  \begin{tabular}{@{}c}\includegraphics[width=0.125\linewidth,height=0.1\linewidth,keepaspectratio,]{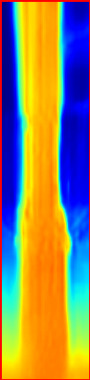}\end{tabular}\hspace{1.5mm}
		  & 
    		\begin{tabular}{@{}c}\includegraphics[width=0.2\linewidth,height=0.1\linewidth,keepaspectratio,]{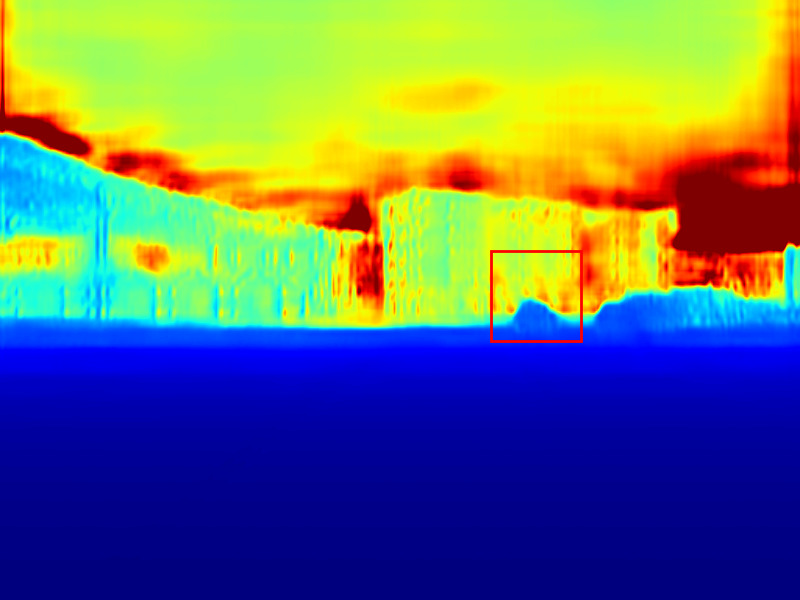}\end{tabular}
		  & 
		  \begin{tabular}{@{}c}\includegraphics[width=0.125\linewidth,height=0.1\linewidth,keepaspectratio,]{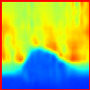}\end{tabular}\hspace{0.75mm}
		  \\
		  
		  &
		             	\begin{tabular}{@{}c}\rotatebox{90}{Ours}\end{tabular}
		  & 
            	\begin{tabular}{@{}c}\includegraphics[width=0.2\linewidth,height=0.1\linewidth,keepaspectratio,]{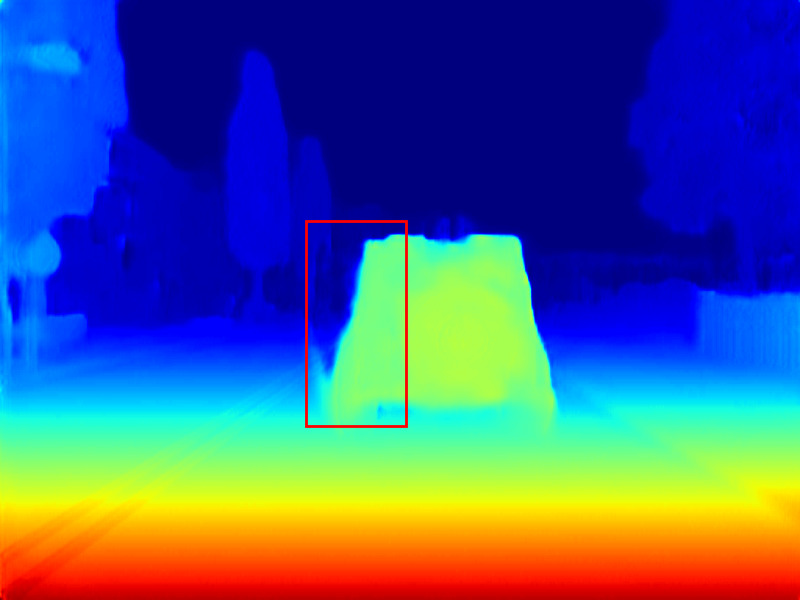}\end{tabular}\hspace{0mm}\hfill
		  &
		  \begin{tabular}{@{}c}\includegraphics[width=0.125\linewidth,height=0.1\linewidth,keepaspectratio,]{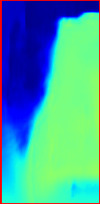}\end{tabular}\hspace{1.5mm}
		  & 
          	\begin{tabular}{@{}c}\includegraphics[width=0.2\linewidth,height=0.1\linewidth,keepaspectratio,]{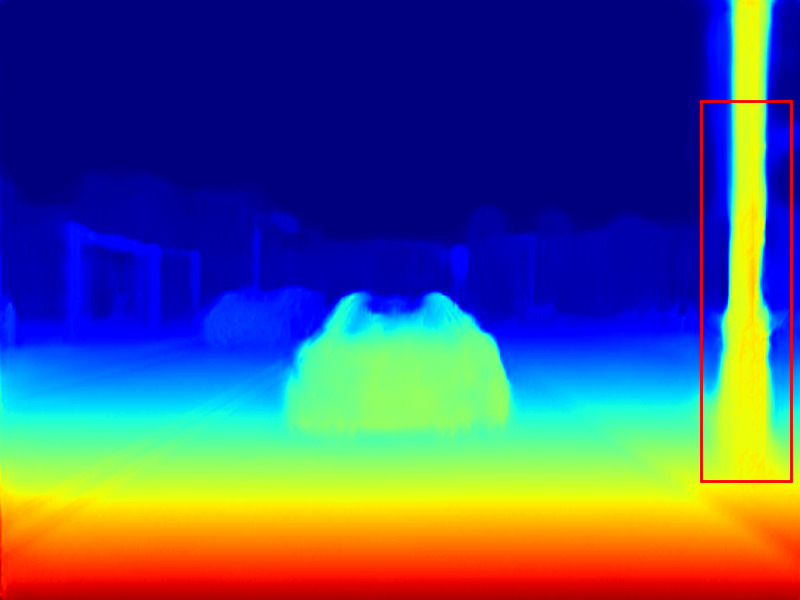}\end{tabular}
		  & 
		  \begin{tabular}{@{}c}\includegraphics[width=0.125\linewidth,height=0.1\linewidth,keepaspectratio,]{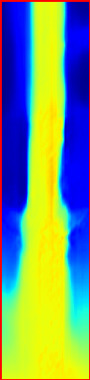}\end{tabular}\hspace{1.5mm}
		  & 
    		\begin{tabular}{@{}c}\includegraphics[width=0.2\linewidth,height=0.1\linewidth,keepaspectratio,]{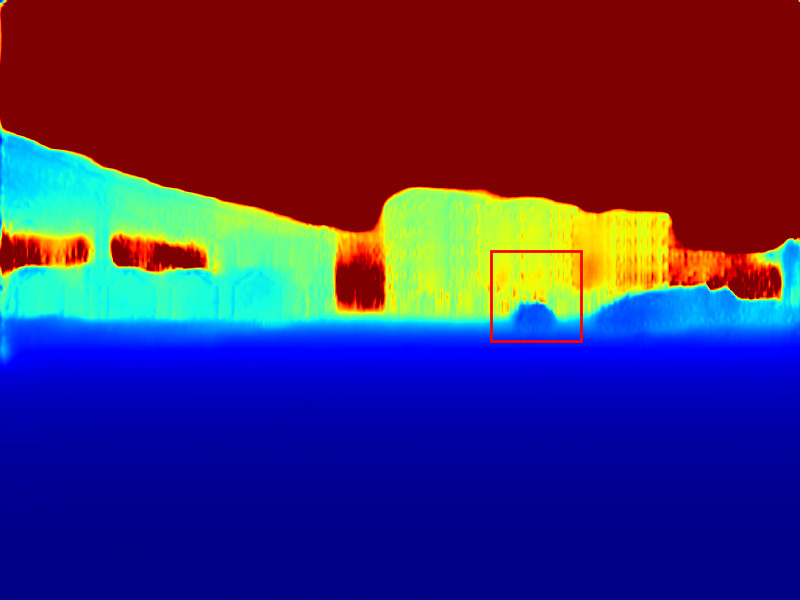}\end{tabular}
		  & 
		  \begin{tabular}{@{}c}\includegraphics[width=0.125\linewidth,height=0.1\linewidth,keepaspectratio,]{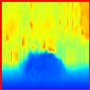}\end{tabular}\hspace{0.75mm}
		  \\
	\end{tabular}
	 \caption{Qualitative comparison of disparity maps (first two columns) and depth maps (last column) on CARLA dataset. Zoomed-in views show that our method produces better results for both close and far objects.}
\label{fig:disp_depth comparison}
	\end{figure*}
	
			\begin{figure*}[t!]
\tiny
     \centering
     \setlength{\tabcolsep}{0mm}
    \begin{tabular}{cccccccc}
        &
          	\begin{tabular}{@{}c}\rotatebox{90}{Input}\end{tabular}
		  & 
            	\begin{tabular}{@{}c}\includegraphics[width=0.2\linewidth,height=0.1\linewidth,keepaspectratio,]{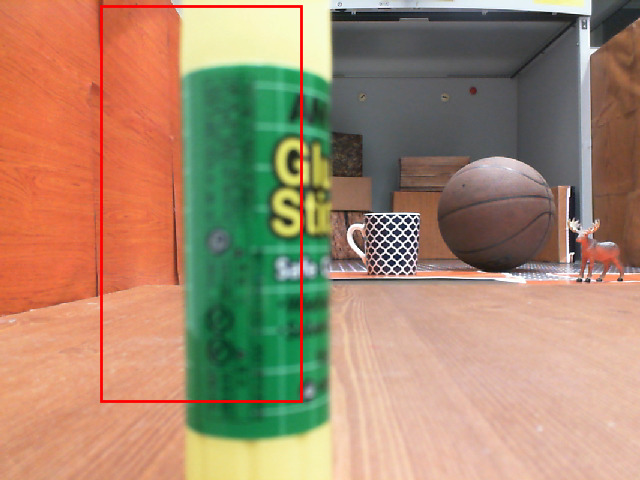}\end{tabular}\hspace{0mm}\hfill
		  &
		  \begin{tabular}{@{}c}\includegraphics[width=0.125\linewidth,height=0.1\linewidth,keepaspectratio,]{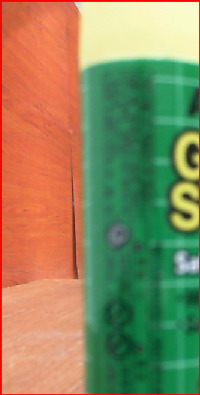}\end{tabular}\hspace{1.5mm}
		  & 
          	\begin{tabular}{@{}c}\includegraphics[width=0.2\linewidth,height=0.1\linewidth,keepaspectratio,]{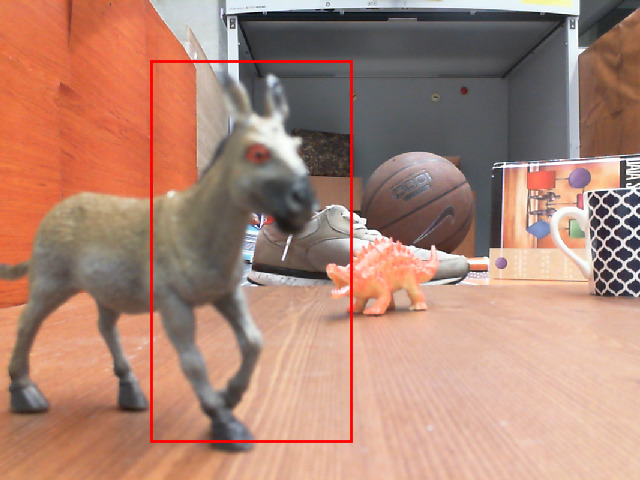}\end{tabular}
		  & 
		  \begin{tabular}{@{}c}\includegraphics[width=0.125\linewidth,height=0.1\linewidth,keepaspectratio,]{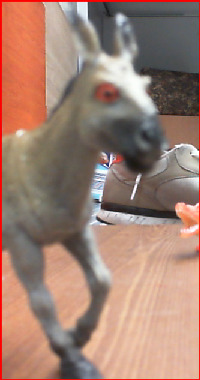}\end{tabular}\hspace{1.5mm}
		  & 
    		\begin{tabular}{@{}c}\includegraphics[width=0.2\linewidth,height=0.1\linewidth,keepaspectratio,]{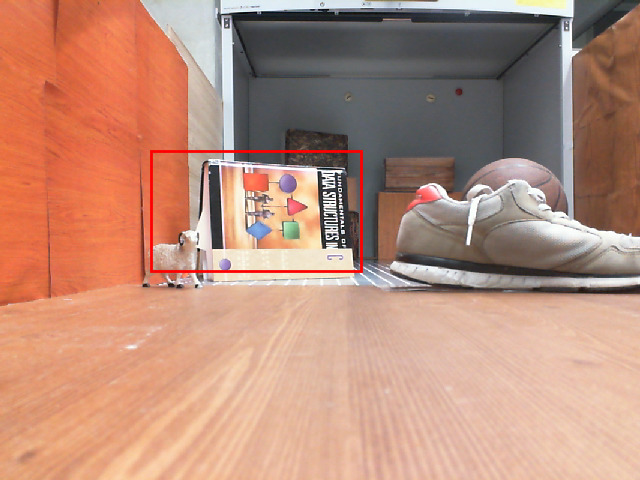}\end{tabular}
		  & 
		  \begin{tabular}{@{}c}\includegraphics[width=0.125\linewidth,height=0.1\linewidth,keepaspectratio,]{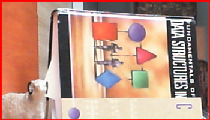}\end{tabular}\hspace{0.75mm}
		  \\
		  
		             	\begin{tabular}{@{}c}\rotatebox{90}{3net}\end{tabular}
		  & 
		  	\begin{tabular}{@{}c}\rotatebox{90}{(54cm) \cite{poggi2018learning}}\end{tabular}
		  	&
            	\begin{tabular}{@{}c}\includegraphics[width=0.2\linewidth,height=0.1\linewidth,keepaspectratio,]{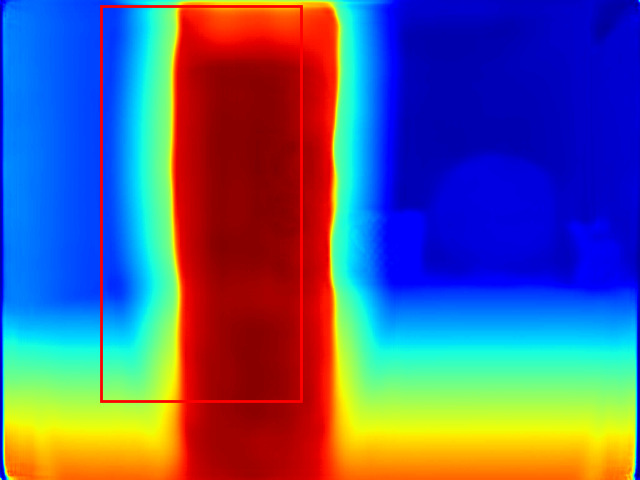}\end{tabular}\hspace{0mm}\hfill
		  &
		  \begin{tabular}{@{}c}\includegraphics[width=0.125\linewidth,height=0.1\linewidth,keepaspectratio,]{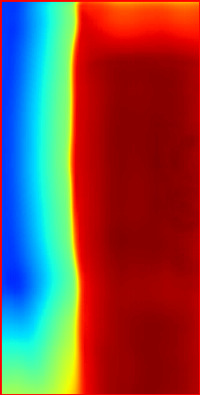}\end{tabular}\hspace{1.5mm}
		  & 
          	\begin{tabular}{@{}c}\includegraphics[width=0.2\linewidth,height=0.1\linewidth,keepaspectratio,]{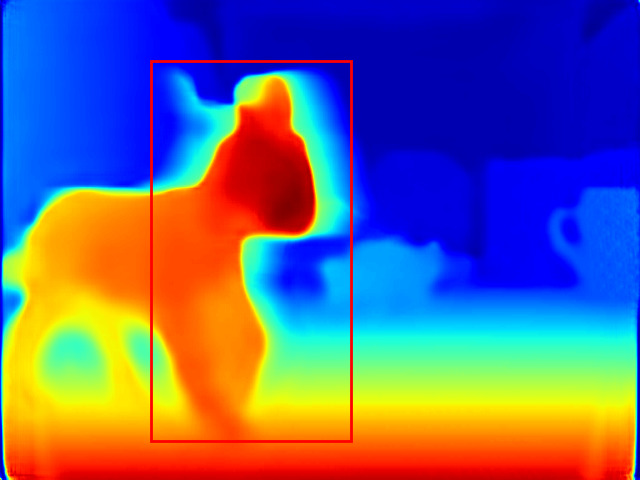}\end{tabular}
		  & 
		  \begin{tabular}{@{}c}\includegraphics[width=0.125\linewidth,height=0.1\linewidth,keepaspectratio,]{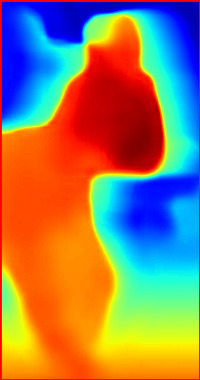}\end{tabular}\hspace{1.5mm}
		  & 
    		\begin{tabular}{@{}c}\includegraphics[width=0.2\linewidth,height=0.1\linewidth,keepaspectratio,]{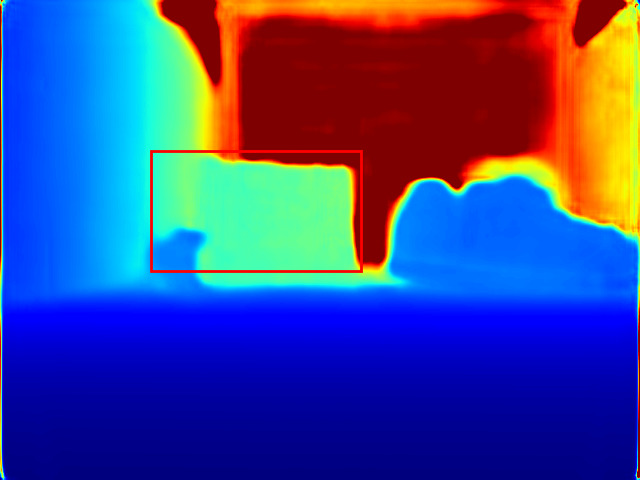}\end{tabular}
		  & 
		  \begin{tabular}{@{}c}\includegraphics[width=0.125\linewidth,height=0.1\linewidth,keepaspectratio,]{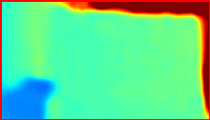}\end{tabular}\hspace{0.75mm}
		  \\
		  
		             	\begin{tabular}{@{}c}\rotatebox{90}{3net}\end{tabular}
		  & 
		  	\begin{tabular}{@{}c}\rotatebox{90}{(10cm) \cite{poggi2018learning}}\end{tabular}
		  & 
            	\begin{tabular}{@{}c}\includegraphics[width=0.2\linewidth,height=0.1\linewidth,keepaspectratio,]{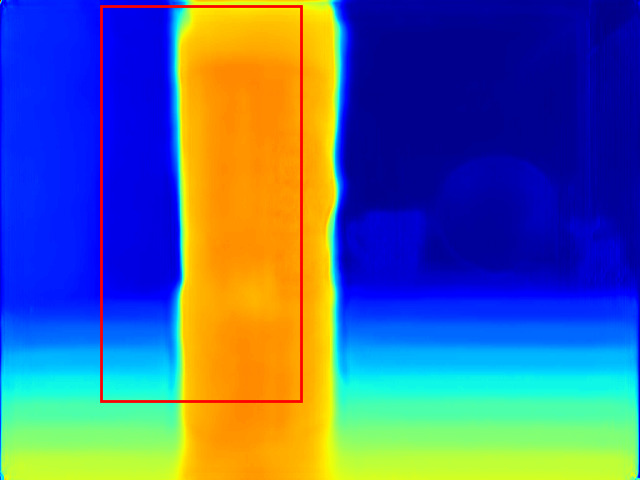}\end{tabular}\hspace{0mm}\hfill
		  &
		  \begin{tabular}{@{}c}\includegraphics[width=0.125\linewidth,height=0.1\linewidth,keepaspectratio,]{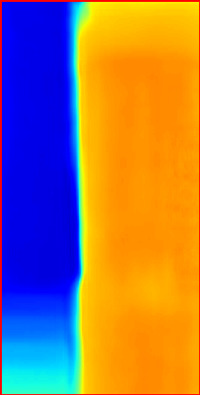}\end{tabular}\hspace{1.5mm}
		  & 
          	\begin{tabular}{@{}c}\includegraphics[width=0.2\linewidth,height=0.1\linewidth,keepaspectratio,]{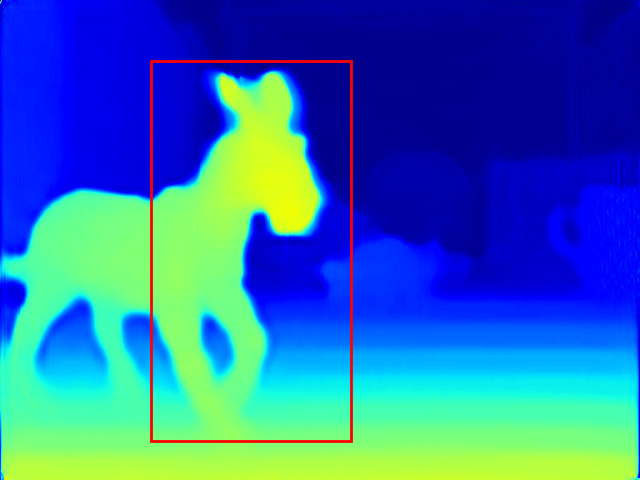}\end{tabular}
		  & 
		  \begin{tabular}{@{}c}\includegraphics[width=0.125\linewidth,height=0.1\linewidth,keepaspectratio,]{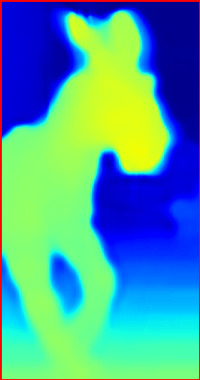}\end{tabular}\hspace{1.5mm}
		  & 
    		\begin{tabular}{@{}c}\includegraphics[width=0.2\linewidth,height=0.1\linewidth,keepaspectratio,]{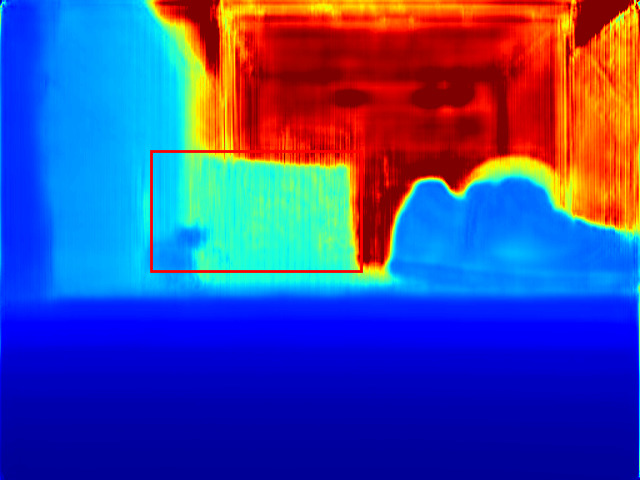}\end{tabular}
		  & 
		  \begin{tabular}{@{}c}\includegraphics[width=0.125\linewidth,height=0.1\linewidth,keepaspectratio,]{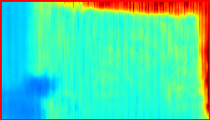}\end{tabular}\hspace{0.75mm}
		  \\
		  
		             	\begin{tabular}{@{}c}\rotatebox{90}{Monodepth}\end{tabular}
		  & 
		  	\begin{tabular}{@{}c}\rotatebox{90}{(54cm) \cite{godard2017unsupervised}}\end{tabular}
		  & 
            	\begin{tabular}{@{}c}\includegraphics[width=0.2\linewidth,height=0.1\linewidth,keepaspectratio,]{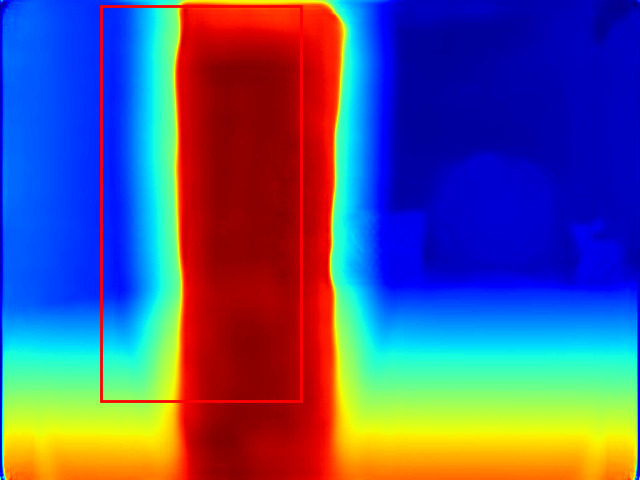}\end{tabular}\hspace{0mm}\hfill
		  &
		  \begin{tabular}{@{}c}\includegraphics[width=0.125\linewidth,height=0.1\linewidth,keepaspectratio,]{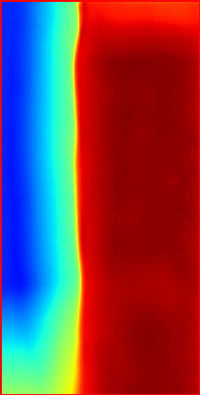}\end{tabular}\hspace{1.5mm}
		  & 
          	\begin{tabular}{@{}c}\includegraphics[width=0.2\linewidth,height=0.1\linewidth,keepaspectratio,]{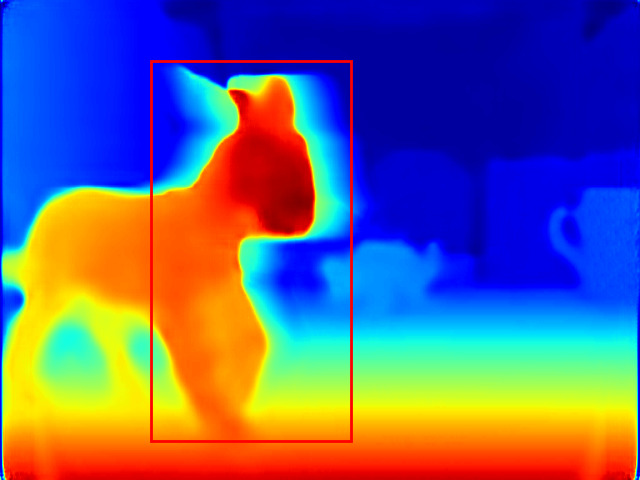}\end{tabular}
		  & 
		  \begin{tabular}{@{}c}\includegraphics[width=0.125\linewidth,height=0.1\linewidth,keepaspectratio,]{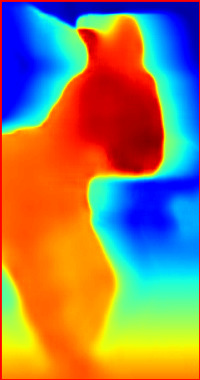}\end{tabular}\hspace{1.5mm}
		  & 
    		\begin{tabular}{@{}c}\includegraphics[width=0.2\linewidth,height=0.1\linewidth,keepaspectratio,]{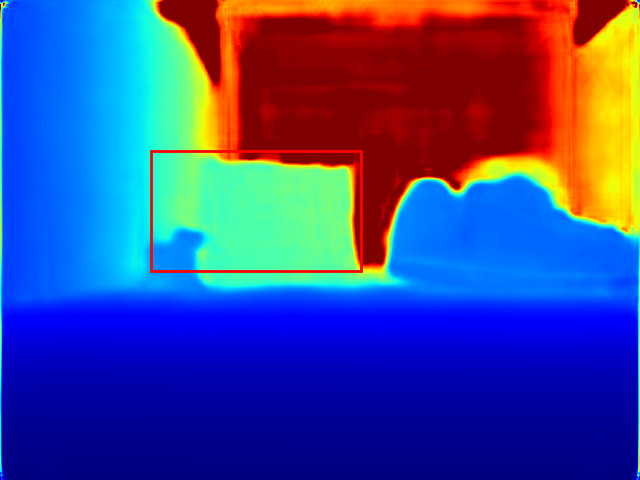}\end{tabular}
		  & 
		  \begin{tabular}{@{}c}\includegraphics[width=0.125\linewidth,height=0.1\linewidth,keepaspectratio,]{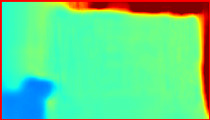}\end{tabular}\hspace{0.75mm}
		  \\
		  
		             	\begin{tabular}{@{}c}\rotatebox{90}{Monodepth}\end{tabular}
		  & 
		  \begin{tabular}{@{}c}\rotatebox{90}{(10cm) \cite{godard2017unsupervised}}\end{tabular}
		  & 
            	\begin{tabular}{@{}c}\includegraphics[width=0.2\linewidth,height=0.1\linewidth,keepaspectratio,]{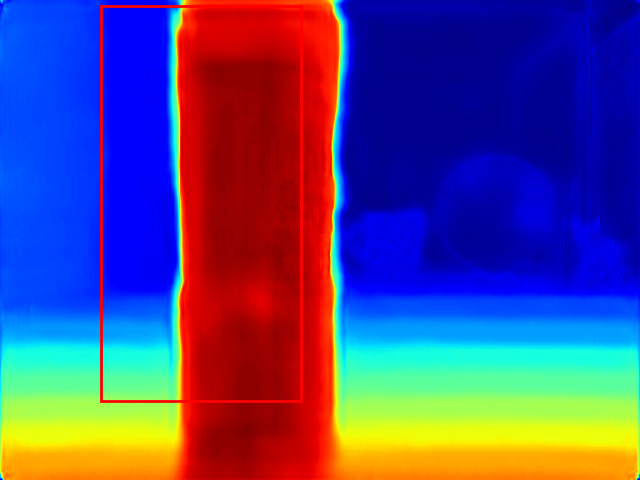}\end{tabular}\hspace{0mm}\hfill
		  &
		  \begin{tabular}{@{}c}\includegraphics[width=0.125\linewidth,height=0.1\linewidth,keepaspectratio,]{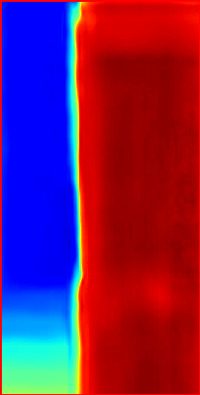}\end{tabular}\hspace{1.5mm}
		  & 
          	\begin{tabular}{@{}c}\includegraphics[width=0.2\linewidth,height=0.1\linewidth,keepaspectratio,]{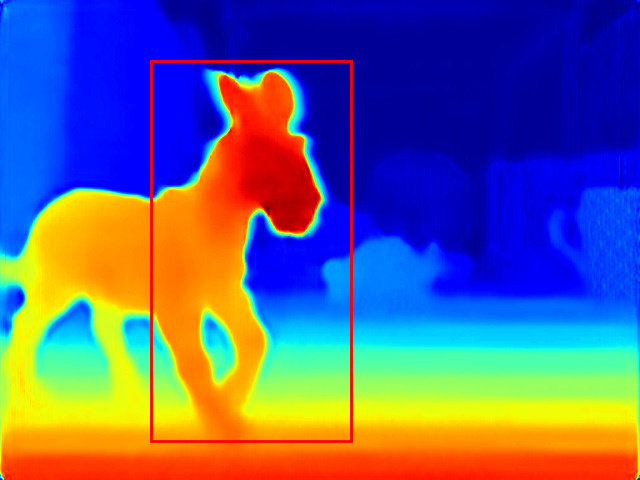}\end{tabular}
		  & 
		  \begin{tabular}{@{}c}\includegraphics[width=0.125\linewidth,height=0.1\linewidth,keepaspectratio,]{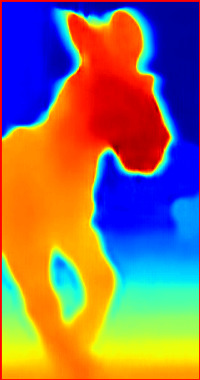}\end{tabular}\hspace{1.5mm}
		  & 
    		\begin{tabular}{@{}c}\includegraphics[width=0.2\linewidth,height=0.1\linewidth,keepaspectratio,]{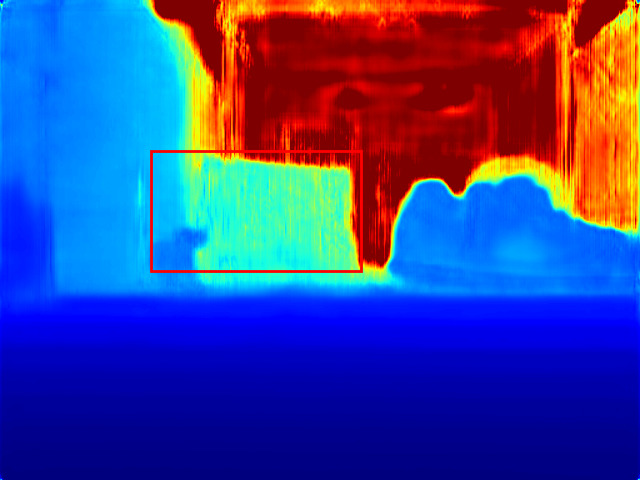}\end{tabular}
		  & 
		  \begin{tabular}{@{}c}\includegraphics[width=0.125\linewidth,height=0.1\linewidth,keepaspectratio,]{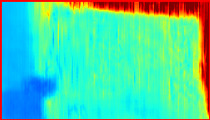}\end{tabular}\hspace{0.75mm}
		  \\

		              	\begin{tabular}{@{}c}\rotatebox{90}{monoResMatch}\end{tabular}
		  & 
		  \begin{tabular}{@{}c}\rotatebox{90}{(54cm) \cite{tosi2019learning}}\end{tabular}
		  & 
            	\begin{tabular}{@{}c}\includegraphics[width=0.2\linewidth,height=0.1\linewidth,keepaspectratio,]{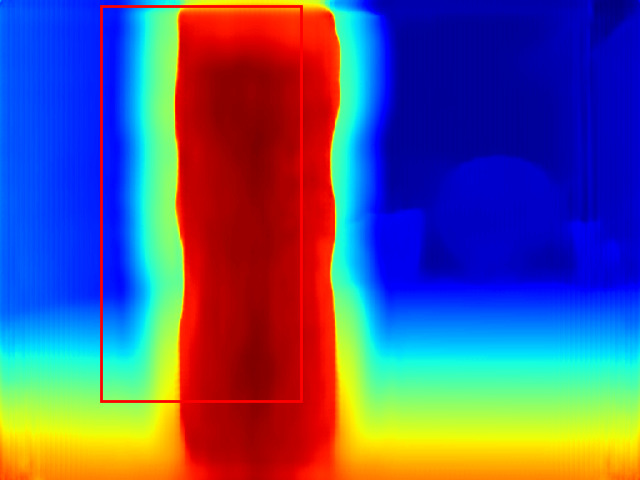}\end{tabular}\hspace{0mm}\hfill
		  &
		  \begin{tabular}{@{}c}\includegraphics[width=0.125\linewidth,height=0.1\linewidth,keepaspectratio,]{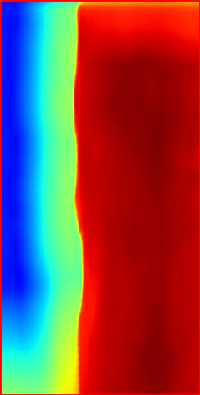}\end{tabular}\hspace{1.5mm}
		  & 
          	\begin{tabular}{@{}c}\includegraphics[width=0.2\linewidth,height=0.1\linewidth,keepaspectratio,]{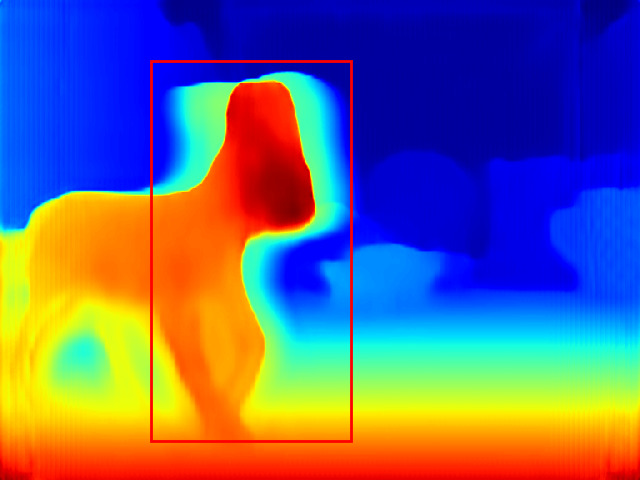}\end{tabular}
		  & 
		  \begin{tabular}{@{}c}\includegraphics[width=0.125\linewidth,height=0.1\linewidth,keepaspectratio,]{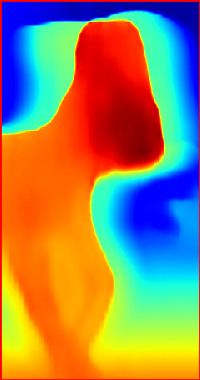}\end{tabular}\hspace{1.5mm}
		  & 
    		\begin{tabular}{@{}c}\includegraphics[width=0.2\linewidth,height=0.1\linewidth,keepaspectratio,]{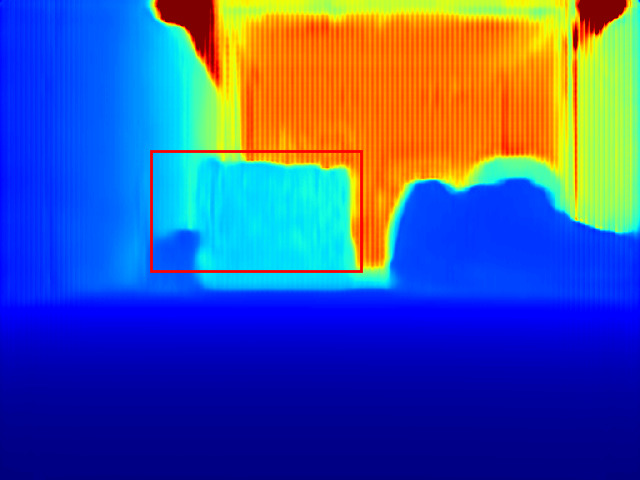}\end{tabular}
		  & 
		  \begin{tabular}{@{}c}\includegraphics[width=0.125\linewidth,height=0.1\linewidth,keepaspectratio,]{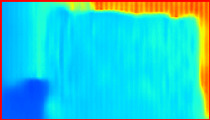}\end{tabular}\hspace{0.75mm}
		  \\
		  
		  \begin{tabular}{@{}c}\rotatebox{90}{monoResMatch}\end{tabular}
		  & 
		           	\begin{tabular}{@{}c}\rotatebox{90}{(10cm) \cite{tosi2019learning}}\end{tabular}
		  & 
            	\begin{tabular}{@{}c}\includegraphics[width=0.2\linewidth,height=0.1\linewidth,keepaspectratio,]{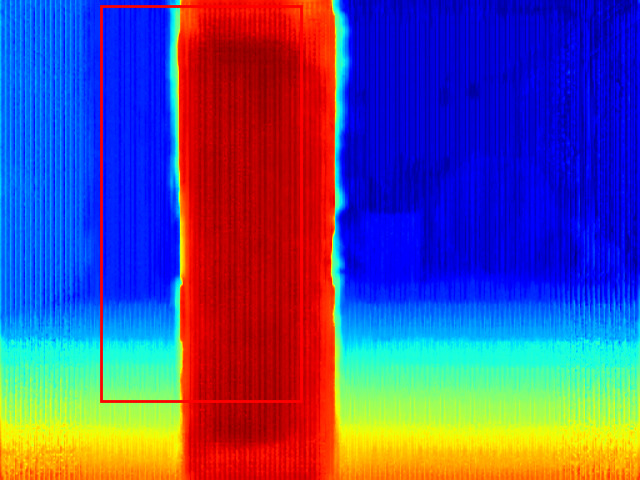}\end{tabular}\hspace{0mm}\hfill
		  &
		  \begin{tabular}{@{}c}\includegraphics[width=0.125\linewidth,height=0.1\linewidth,keepaspectratio,]{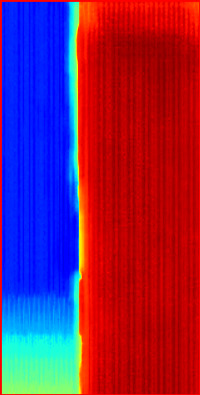}\end{tabular}\hspace{1.5mm}
		  & 
          	\begin{tabular}{@{}c}\includegraphics[width=0.2\linewidth,height=0.1\linewidth,keepaspectratio,]{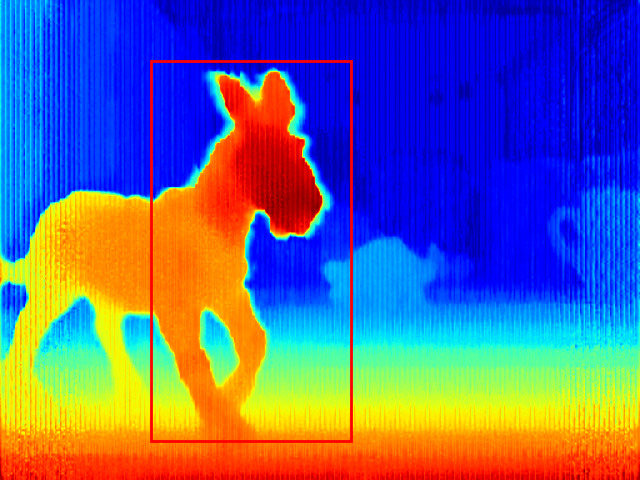}\end{tabular}
		  & 
		  \begin{tabular}{@{}c}\includegraphics[width=0.125\linewidth,height=0.1\linewidth,keepaspectratio,]{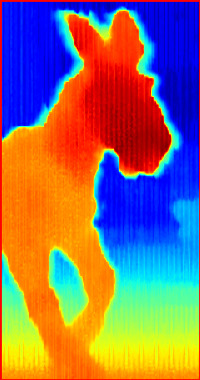}\end{tabular}\hspace{1.5mm}
		  & 
    		\begin{tabular}{@{}c}\includegraphics[width=0.2\linewidth,height=0.1\linewidth,keepaspectratio,]{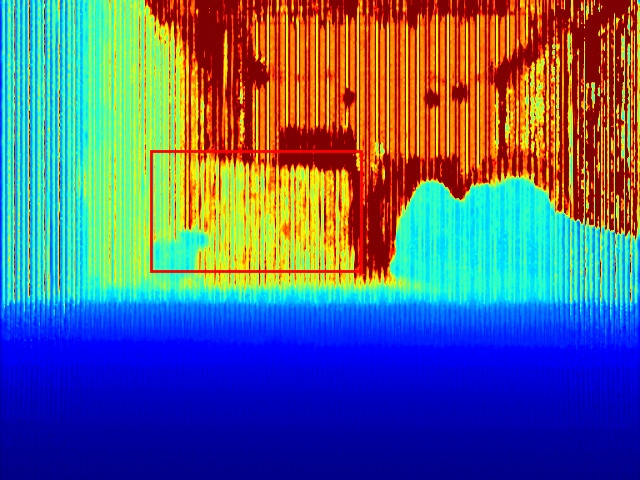}\end{tabular}
		  & 
		  \begin{tabular}{@{}c}\includegraphics[width=0.125\linewidth,height=0.1\linewidth,keepaspectratio,]{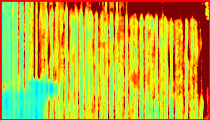}\end{tabular}\hspace{0.75mm}
		  \\
		  		  		             	\begin{tabular}{@{}c}\rotatebox{90}{Monodepth2}\end{tabular}
		  & 
		  \begin{tabular}{@{}c}\rotatebox{90}{(54cm) \cite{godard2019digging}}\end{tabular}
		  & 
            	\begin{tabular}{@{}c}\includegraphics[width=0.2\linewidth,height=0.1\linewidth,keepaspectratio,]{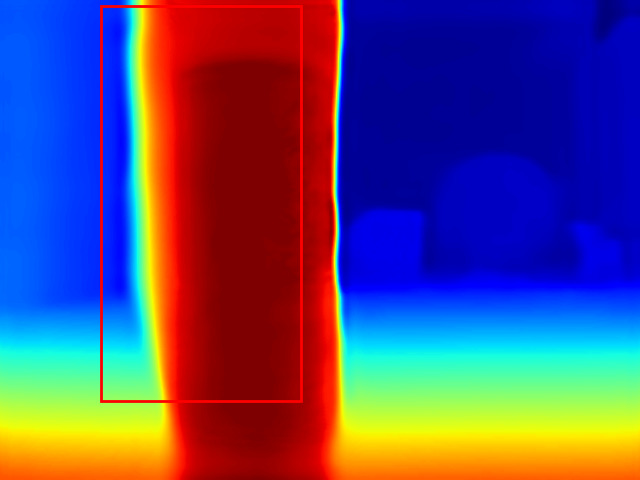}\end{tabular}\hspace{0mm}\hfill
		  &
		  \begin{tabular}{@{}c}\includegraphics[width=0.125\linewidth,height=0.1\linewidth,keepaspectratio,]{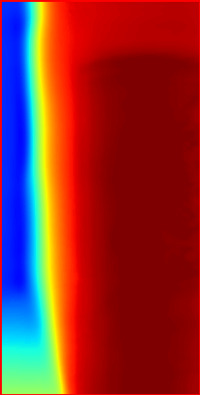}\end{tabular}\hspace{1.5mm}
		  & 
          	\begin{tabular}{@{}c}\includegraphics[width=0.2\linewidth,height=0.1\linewidth,keepaspectratio,]{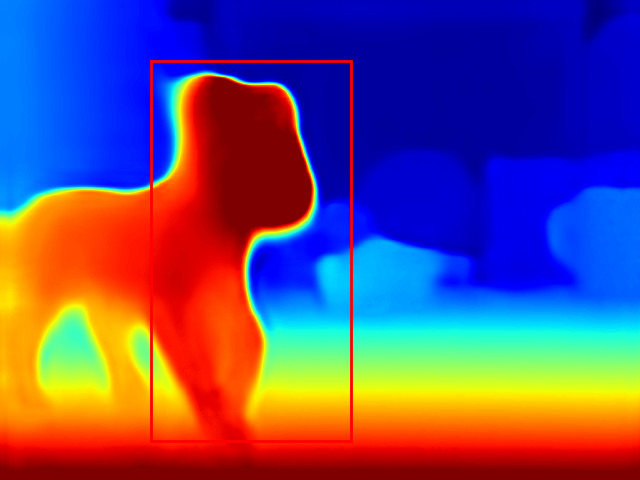}\end{tabular}
		  & 
		  \begin{tabular}{@{}c}\includegraphics[width=0.125\linewidth,height=0.1\linewidth,keepaspectratio,]{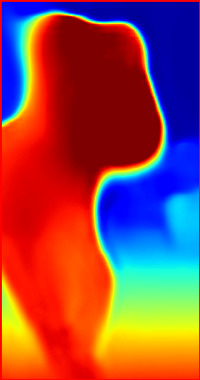}\end{tabular}\hspace{1.5mm}
		  & 
    		\begin{tabular}{@{}c}\includegraphics[width=0.2\linewidth,height=0.1\linewidth,keepaspectratio,]{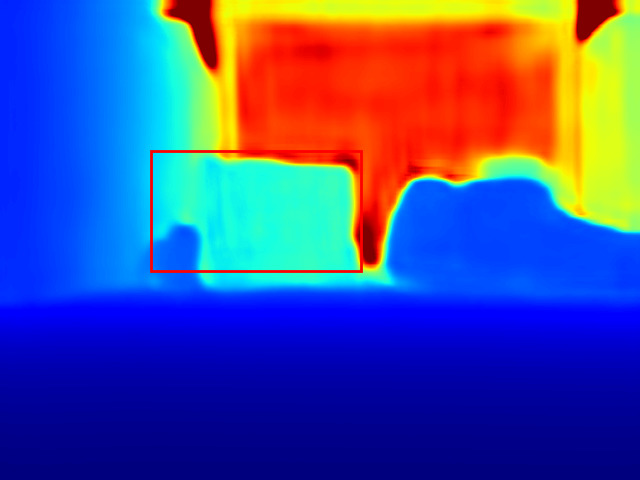}\end{tabular}
		  & 
		  \begin{tabular}{@{}c}\includegraphics[width=0.125\linewidth,height=0.1\linewidth,keepaspectratio,]{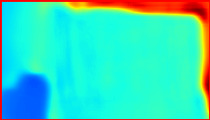}\end{tabular}\hspace{0.75mm}
		  \\
		  		             	\begin{tabular}{@{}c}\rotatebox{90}{Monodepth2}\end{tabular}
		  & 
		  \begin{tabular}{@{}c}\rotatebox{90}{(10cm) \cite{godard2019digging}}\end{tabular}
		  & 
            	\begin{tabular}{@{}c}\includegraphics[width=0.2\linewidth,height=0.1\linewidth,keepaspectratio,]{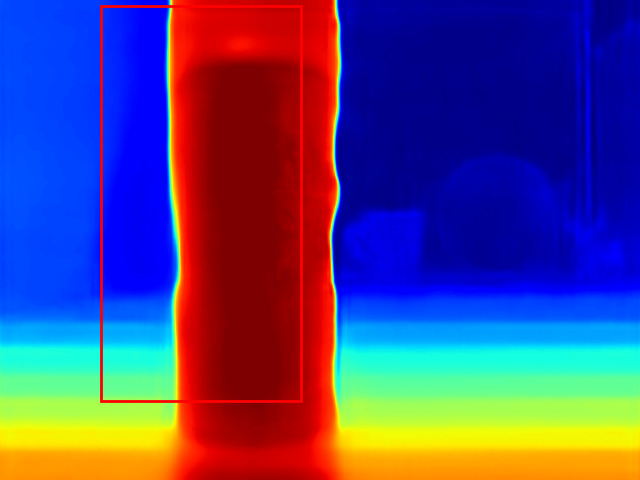}\end{tabular}\hspace{0mm}\hfill
		  &
		  \begin{tabular}{@{}c}\includegraphics[width=0.125\linewidth,height=0.1\linewidth,keepaspectratio,]{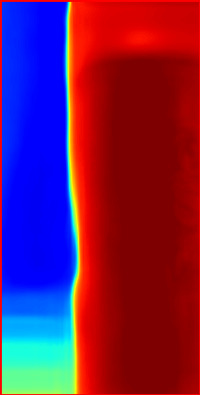}\end{tabular}\hspace{1.5mm}
		  & 
          	\begin{tabular}{@{}c}\includegraphics[width=0.2\linewidth,height=0.1\linewidth,keepaspectratio,]{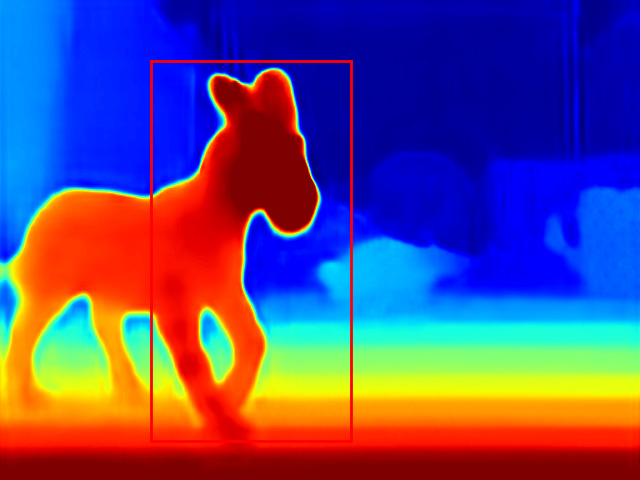}\end{tabular}
		  & 
		  \begin{tabular}{@{}c}\includegraphics[width=0.125\linewidth,height=0.1\linewidth,keepaspectratio,]{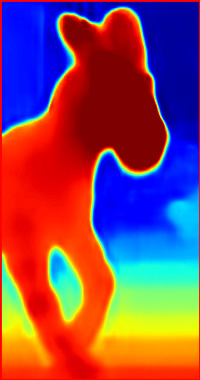}\end{tabular}\hspace{1.5mm}
		  & 
    		\begin{tabular}{@{}c}\includegraphics[width=0.2\linewidth,height=0.1\linewidth,keepaspectratio,]{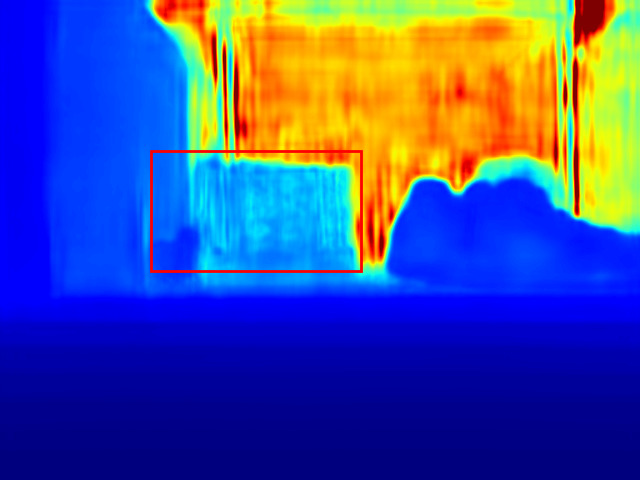}\end{tabular}
		  & 
		  \begin{tabular}{@{}c}\includegraphics[width=0.125\linewidth,height=0.1\linewidth,keepaspectratio,]{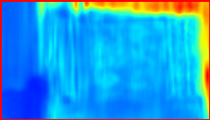}\end{tabular}\hspace{0.75mm}
		  \\

		  &
		             	\begin{tabular}{@{}c}\rotatebox{90}{Ours}\end{tabular}
		  & 
            	\begin{tabular}{@{}c}\includegraphics[width=0.2\linewidth,height=0.1\linewidth,keepaspectratio,]{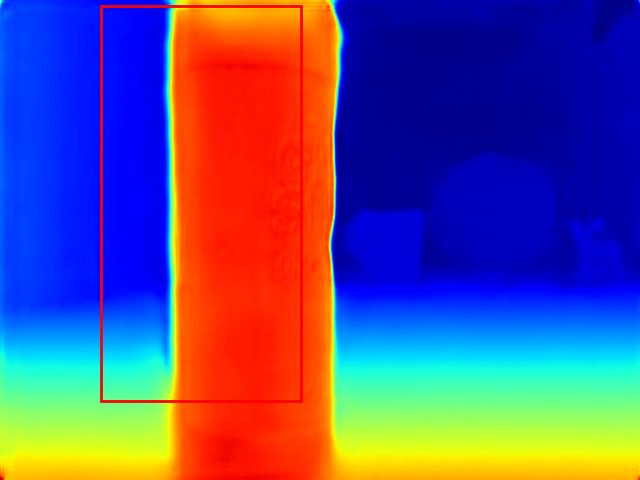}\end{tabular}\hspace{0mm}\hfill
		  &
		  \begin{tabular}{@{}c}\includegraphics[width=0.125\linewidth,height=0.1\linewidth,keepaspectratio,]{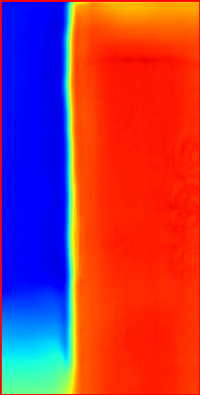}\end{tabular}\hspace{1.5mm}
		  & 
          	\begin{tabular}{@{}c}\includegraphics[width=0.2\linewidth,height=0.1\linewidth,keepaspectratio,]{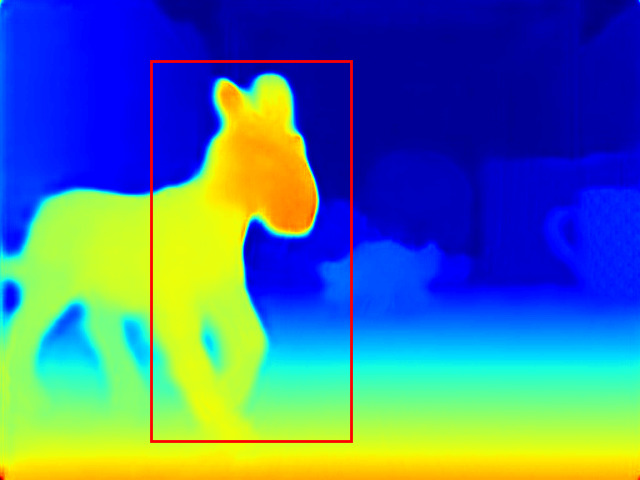}\end{tabular}
		  & 
		  \begin{tabular}{@{}c}\includegraphics[width=0.125\linewidth,height=0.1\linewidth,keepaspectratio,]{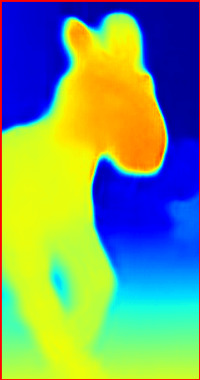}\end{tabular}\hspace{1.5mm}
		  & 
    		\begin{tabular}{@{}c}\includegraphics[width=0.2\linewidth,height=0.1\linewidth,keepaspectratio,]{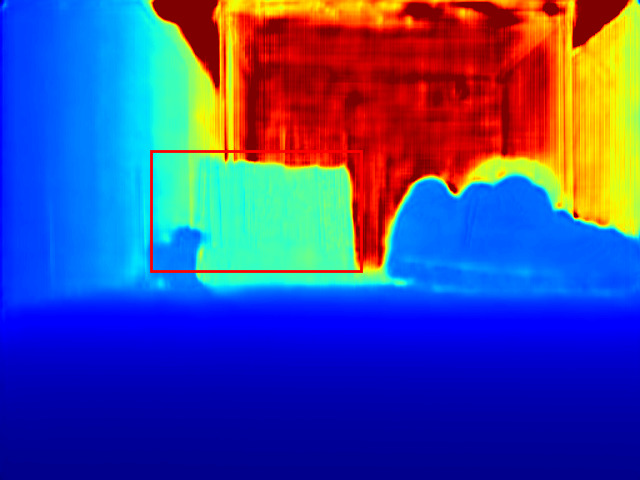}\end{tabular}
		  & 
		  \begin{tabular}{@{}c}\includegraphics[width=0.125\linewidth,height=0.1\linewidth,keepaspectratio,]{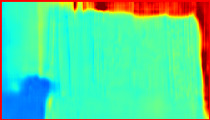}\end{tabular}\hspace{0.75mm}
		  \\
	\end{tabular}
	 \caption{Qualitative comparison of disparity maps (first two columns) and depth maps (last column) on small objects dataset. Our method generates more accurate predictions with sharp boundaries.}
\label{fig:real disp_depth comparison}
	\end{figure*}
\section{Experimental Results}
Due to the unavailability of multi-baseline stereo datasets, researchers have not investigated the advantages of using more than one baseline during training. Although some researchers \cite{shinzato2016carina,maddern20171} have acquired trinocular stereo datasets from Bumblebee XB3 camera, they are more focused on localization and mapping. Moreover, these datasets do not provide the calibration parameters to horizontally align the three views. As the existing stereo-based unsupervised methods require rectified image pairs for training, such datasets are not feasible to use for the task of multi-baseline depth estimation. We develop our own dataset to show the importance of using multiple baselines during training.

We evaluate the effectiveness of our approach both qualitatively and quantitatively. We compare the performance of our approach with the stereo-based methods proposed in the past. We compare our results against Monodepth \cite{godard2017unsupervised}, monoResMatch \cite{tosi2019learning}, Monodepth2 \cite{godard2019digging}, and 3Net \cite{poggi2018learning}. For fair comparison, we train monoResMatch without the proxy-supervised loss. The results validate that the proposed approach yields more accurate disparity predictions.
\subsection{Datasets}
\textbf{CARLA Dataset} CARLA simulator \cite{Dosovitskiy17} is used to acquire the multi-baseline dataset. We attach three cameras to the vehicle in a parallel arrangement to obtain horizontally aligned images. In addition, we also add a depth sensor to get ground truth depth maps for evaluation. We choose the large baseline to be 54 $cm$, which is equal to the baseline used in KITTI dataset \cite{Menze2015CVPR}. The small baseline is chosen as 10 $cm$. The simulator is run in auto-pilot mode under clear weather conditions to gather the dataset. The dataset consists of approximately 14000 images, out of which 1300 images are used for evaluation while the remaining are used for training. 

For evaluation, we use the metrics given in \cite{eigen2014depth}: Abs Rel, Sq Rel, RMSE linear, RMSE log, and threshold-based metrics $\delta$. The predicted disparity maps are converted to depth maps using baseline and focal length to compute these errors. For our approach, we use 54 $cm$ baseline to get depth predictions. We post-processed the results of Monodepth \cite{godard2017unsupervised}, monoResMatch \cite{tosi2019learning}, and 3Net \cite{poggi2018learning} using the method applied in \cite{godard2017unsupervised}. We report the results with and without post-processing.

Table \ref{tab:comp_wopp} shows the detailed results on different depth ranges. From the results, it is seen that Monodepth \cite{godard2017unsupervised}, Monodepth2 \cite{godard2019digging} and 3Net \cite{poggi2018learning} trained on wide baseline perform better at depth range greater than 10 meters. This is due to the fact that wide baseline can not deal with the occlusions caused by near objects. The monoResMatch \cite{tosi2019learning} trained on 10 $cm$ baseline performs worst among all the methods. For 0.1 to 80 meters depth range, our approach outperforms previous methods. This is expected as our method makes use of both narrow and wide baselines. Although post-processing considerably improves the results Monodepth, monoResMatch and 3Net, our approach does not require any post-processing. 

For qualitative comparison, we provide disparity maps and zoomed-in views as shown in Figure \ref{fig:disp_depth comparison}. Results illustrate that small baseline tends to over-smooth the disparity predictions in far regions due to lower depth resolution. On the contrary, wide baseline provides more accurate predictions for far regions but generates severe occlusion artifacts for closer surfaces. From the results, it is obvious that for close and far regions, our approach performs similar to small and large baselines, respectively. Hence, producing much improved depth estimates.

\textbf{Small Objects Dataset} We prepare another dataset to demonstrate the usefulness of our approach on real scenes. The dataset is captured using Microsoft LifeCam webcam. Instead of using three cameras, we employ single camera and displace it laterally to acquire three parallel images of each scene similar to \cite{imran2020unsupervised}. Zaber's A-LSQ600D motorized linear translation stage is used to control the position of camera. We capture the images of small objects to build the dataset of 5800 images. Training and test sets contain 5500 and 300 images respectively. All the objects are placed within two meters range. We set the small and wide baselines to 2 $mm$ and 10 $mm$, respectively.  

We show the results in Figure \ref{fig:real disp_depth comparison}. Results clearly depict the superior performance of our approach over single baseline methods. The disparity maps of Figure \ref{fig:real disp_depth comparison} illustrate that for closer objects, our method produces crisp and accurate disparity maps similar to small baseline. In contrast, wide baseline generates serious artifacts near close object boundaries. The depth maps provide the clear picture of far objects. For far objects, small baseline fails to estimate accurate depth. On the other hand, multi-baseline method gives much accurate depth predictions similar to wide baseline. Again the results verify the efficacy of multi-baseline training over single baseline training.

\subsection{Implementation Details} We implement our network in Tensorflow \cite{tensorflow2015-whitepaper}. We train the model for 70 epochs with batch size of 8. All other hyperparameters are set as in \cite{godard2017unsupervised}. We use Nvidia GTX 1080 GPU for experiments. Training on 13000 images for 70 epochs takes around 36 hours. The number of trainable parameters are approximately 66.2 million. It should be noted that at test time, we only use the output of decoder 3; therefore, depth prediction is fast and inference time of the network is 21 frames per second. A CPU implementation on an Intel Core i5 processor with 8GB RAM provides 2 frames per second.

\section{Conclusion}
In this work, we propose a novel multi-baseline technique for unsupervised monocular depth estimation. We overcome the shortcomings of single-baseline stereo supervision by training the model with two stereo baselines. Our model combines the advantages of small and wide baseline stereo systems. Unlike previous stereo approaches that work well in a certain range, our method generates accurate disparity maps both in near and far ranges. Furthermore, our method uses only a single camera at test time to predict multi-baseline depth. This is in contrast to traditional multi-baseline systems, which require more than two cameras to provide real time depth. Therefore, the proposed method is well-suited for practical applications.

{\small
\bibliographystyle{IEEEtran}
\bibliography{ref}
}




\end{document}